\documentclass[11pt]{article}
\usepackage{arxiv}
\usepackage{graphicx}
\graphicspath{{figures/}}
\usepackage{subcaption}
\usepackage{pgfplots}
\pgfplotsset{compat=1.18}
\usetikzlibrary{patterns}

\title{PreAct: Computer-Using Agents that Get\\Faster on Repeated Tasks}

\author{
  Bojie Li \\
  Pine AI
}

\date{}
\runningtitle{PreAct: Computer-Using Agents that Get Faster on Repeated Tasks}

\begin{document}

\maketitle

\begin{abstract}
Computer-using agents drive real software through the screen---clicking and typing---but they solve every task from scratch: asked to repeat a task, an agent re-reads the screen, re-reasons every tap, and pays the full cost again. We present PreAct, which lets such an agent get faster on tasks it has done before. The first time it succeeds, PreAct compiles the run into a small state-machine program---states that check the screen, transitions that act---and on later runs replays it directly instead of invoking the agent---$8.5$--$13\times$ faster, with no per-step language-model calls. Replay is not blind: at each step PreAct checks that the screen matches what the program expects before acting, and hands control back to the agent the moment something is off.

PreAct applies the same discipline when deciding what to keep: a freshly compiled program enters the store only if, re-run from a clean state, an independent evaluator confirms it solved the task---catching programs that replay to their last step yet leave the task undone. Across a mobile, a desktop, and a web benchmark, this store-time check separates repeated runs that improve from ones that degrade as faulty programs accumulate---worth $1.75$--$2.6$ tasks per benchmark, the same direction on all three; a fallback that explores afresh when no program fits brings PreAct level with a strong record-and-replay baseline. We also report what did not matter: prompt wording, runtime guardrails, and whether a language model or a plain embedding retriever selects which program to reuse.
\end{abstract}

\begin{center}
\small
Code: \url{https://github.com/19PINE-AI/PreAct} \\[2pt]
Website: \url{https://01.me/research/PreAct/}
\end{center}
\vspace{-0.6em}

\section{Introduction}
\label{sec:intro}

Computer-using agents---models that operate ordinary software by reading the screen and issuing clicks and keystrokes~\cite{yao2023react,anthropic2024computeruse}---are now good enough to handle everyday chores like booking a meeting or filing an expense. They are also wasteful. Consider a task from the AndroidWorld benchmark~\cite{rawles2024androidworld}: \emph{add a new contact named ``Emilia Gonzalez'' with a given phone number}. The agent reads a screenshot, decides where to tap, acts, and repeats, spending half a dozen rounds of vision-and-language inference to walk through the contacts app. Ask for the same task tomorrow and it starts over. The screen has not changed and neither have the taps, but the agent kept nothing it could simply run again, so it pays the full cost a second time.

\begin{figure}[t]
\centering
\includegraphics[width=0.74\linewidth]{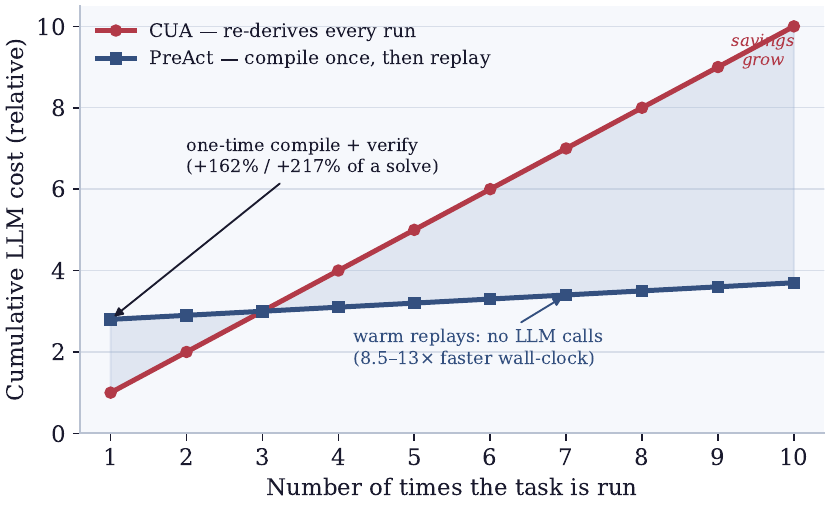}
\caption{The cost of doing the same task twice. A standard computer-using agent re-derives a familiar task on every run, so its cumulative LLM cost grows with the number of runs (red). PreAct pays a one-time premium to compile and check the first success (the bump at run~1), then replays the stored program on every later run with no per-step language-model calls (blue), so its cumulative cost barely rises; replays also run $8.5$--$13\times$ faster in wall-clock time. The more often a task recurs, the more this saves. Cost is normalised to one full agent solve; the wall-clock speedup and the one-time overhead are measured, not illustrative.}
\label{fig:rerun}
\end{figure}

This is not how people work. Someone opening Photoshop for the first time, or a traveler in China reaching for Alipay or WeChat Pay to settle a bill, is painfully slow at first---reading every menu, hunting for the right button. After a few repetitions they are nearly as fluent as a local: the deliberate, perception-heavy reasoning of the first attempt has been internalized into an almost automatic routine. Yet that fluency is not blind---they still glance at the screen and act only when it looks as they expect; the moment the network stalls or an unfamiliar dialog appears, they notice at once and slow back down to think. Fluency is prediction, not blindness. The animating question behind PreAct is whether a computer-using agent can acquire the same continual-learning capability---not merely solving a task it has seen before, but solving it \emph{faster and more cheaply} the second time, much as a practiced human does. This efficiency dimension is largely absent from how agents are evaluated: most benchmarks score only whether a task \emph{succeeds}, not how much perception and reasoning a repeated success costs, so an agent that re-derives every task from scratch looks no worse than one that has learned to skip the deliberation. We treat repeated-task efficiency as a first-class outcome and measure it directly (Figure~\ref{fig:rerun}).

Several lines of work try to reuse past behaviour, but each keeps a model in the execution loop or stores something it cannot reliably re-run. Skill libraries such as SAGE~\cite{liang2024sage} save parameterised behaviours, yet a language model still carries them out at run time, so most of the per-step cost remains. Compilation systems~\cite{compiledai2026,actionengine2026} turn behaviour into deterministic code, but they target narrow business logic, or build a program by crawling an application rather than from an actual successful run. Record-and-replay systems~\cite{musclemem,agentrr2025,workflowuse} cache the sequence of actions from one run and replay it, falling back to the agent when the cache misses; the cached trace has no way to tell whether each step really landed, no room for branching, and no means of improving itself.

\begin{figure}[t]
\centering
\includegraphics[width=0.92\linewidth]{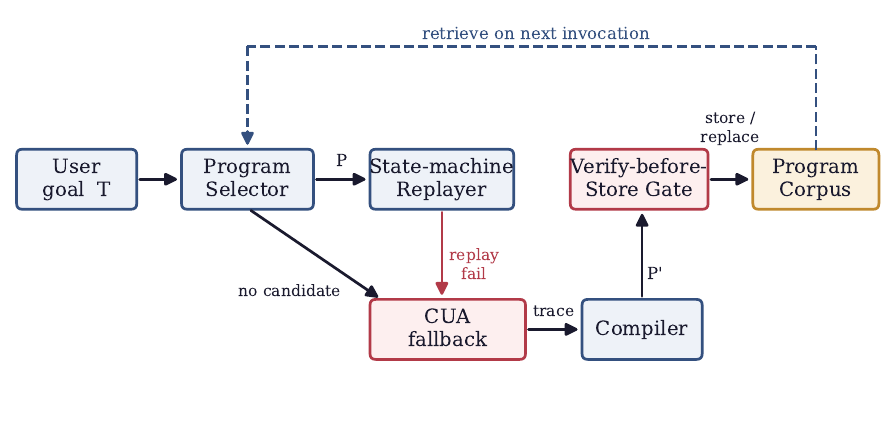}
\caption{The PreAct harness is a verified compile--extend--replace loop. A goal is routed to the \emph{Program Selector}; a retrieved program is run by the \emph{Replayer}, which walks the graph and checks each predicate against the live screen. On any miss the harness falls back to the full agent (\emph{CUA}), whose trace is compiled into a new program $P'$. $P'$ enters the corpus only through the \emph{Verify-before-Store Gate}. The corpus is the only growing structure; the harness code is fixed.}
\label{fig:arch}
\end{figure}

PreAct takes a different position: the thing the agent stores is exactly the thing it later runs. The first time the agent finishes ``add Emilia Gonzalez,'' PreAct turns the run into a small state machine. Each state carries a check on the screen---the contact form is open; the name field now reads the value we typed---and each transition carries an action, such as tapping \textsf{Create contact} or typing the first name. The next time a matching task arrives, PreAct retrieves that state machine and walks it: at every state it confirms the screen looks right before doing anything, and at every transition it performs the action. No language model runs during a successful replay, which is where the order-of-magnitude saving comes from. This observe-first-then-act discipline is what makes direct replay trustworthy and what sets PreAct apart from a recorded macro: where an RPA script or a cached action trace fires its stored steps blindly, PreAct checks the screen before each one. When a check does fail---the app has changed, an unexpected dialog appears---PreAct hands control back to the full agent, which finishes the task and produces a fresh program. Over time the store grows on its own: the code that drives this loop is fixed, and what accumulates is a library of programs the agent can run directly (Figure~\ref{fig:arch}).

The same check runs a second time, at the moment PreAct decides what to remember---and getting this right took us the longest. A compiled program can replay all the way to its final step and still leave the task undone: it taps through the contact form and presses Save, but a stale field meant the name was never entered, so no contact exists. An agent that trusts such a program will fail the same way every time it reuses it, and the more it runs the worse it becomes. PreAct's answer is to apply the run-time discipline once more before storage: right after compiling, it resets the environment, runs the new program from scratch, and asks the benchmark's own evaluator whether the task was actually solved; only then is the program allowed into the store. The same principle thus operates at two timescales---before each action on replay, and before a program is kept---and the second is the difference between a system that gets better with use and one that quietly decays.

We test PreAct on three platforms---a phone (AndroidWorld), a desktop (OSWorld), and a website (WebArena)---where warm replays run $8.5$--$13\times$ faster than re-deriving the task from scratch and make no per-step language-model calls. Turning the check on or off, with a fresh corpus or one the agent has already built, gives a consistent picture: with the check, repeated runs get better; without it, they get worse as faulty programs collect. The gate is worth $1.75$--$2.6$ tasks across the cold$\to$warm transition, pointing the same way on every platform. A second, smaller idea matters too---when no stored program fits a task, exploring it afresh rather than giving up keeps PreAct level with a strong record-and-replay baseline (Muscle-Mem). We are also careful to report what did \emph{not} change the numbers, since it is easy to credit the wrong thing: the exact wording of the prompts, a handful of hand-written runtime guardrails, and whether a language model or a plain embedding retriever (which matches the agentic selector at $100\%$ vs.\ $75.6\%$ retrieval) picks which stored program to reuse all made little or no difference. What carries the result is the design itself---store the program you run, and never keep one you have not checked.

\section{Related Work}
\label{sec:related}

\begin{figure}[t]
\centering
\includegraphics[width=0.66\linewidth]{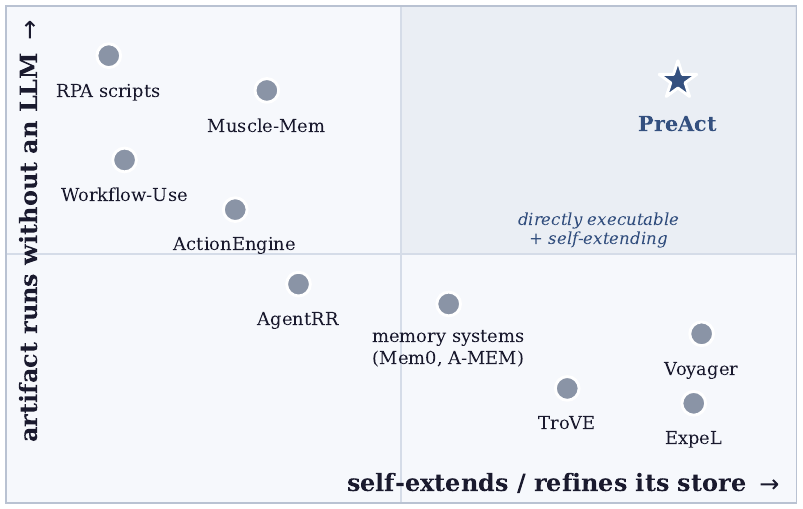}
\caption{Where PreAct sits, along the two properties it is designed to combine. Many prior systems achieve one or the other. \emph{Record-and-replay} systems (Muscle-Mem, Workflow-Use) and classical RPA run their stored artifact \emph{without} an LLM, but only append it (top-left). \emph{Skill and memory} systems---Voyager, TroVE, ExpeL, Mem0/A-MEM---genuinely self-extend, and some (Voyager, TroVE) even self-verify what they add, but they invoke it \emph{through} an LLM at run time (bottom-right). PreAct's distinction is the \emph{combination}: a directly-executable artifact (no LLM in the loop at replay) grown through verified, in-place refinement. Axes are qualitative.}
\label{fig:landscape}
\end{figure}

PreAct sits at the intersection of six literatures: computer-using agents, the reason--act loop they run, skill and memory systems that try to reuse past behaviour, code as an agent representation, world models that learn environment dynamics, and systems that compile or replay workflows. We survey each in turn, then (Table~\ref{tab:related}, Figure~\ref{fig:landscape}) locate PreAct relative to the closest prior work.

\begin{table}[t]
\centering
\caption{PreAct vs.\ related approaches.}
\label{tab:related}
\footnotesize
\setlength{\tabcolsep}{4pt}
\begin{tabular}{@{}p{1.7cm}p{3.0cm}p{2.5cm}p{3.0cm}p{3.7cm}@{}}
\toprule
\textbf{System} & \textbf{Memory representation} & \textbf{Replay fidelity} & \textbf{Fallback path} & \textbf{Self-extension} \\
\midrule
Skill libraries (e.g.\ Voyager) & Code skills & LLM-bound (skill called by agent) & LLM & Grows/self-verifies skill library \\
Muscle-Mem & Linear tool calls & Direct & Agent on cache miss & Append cache only \\
AgentRR & Multi-level experiences & Indirect & LLM & Append only \\
Workflow-Use & Linear scripts & Direct & Agent & Append only \\
ActionEngine & State machine via crawl & Flat-Python generated & None & Re-crawl from scratch \\
\textbf{PreAct} & \textbf{State machine} & \textbf{Runs the state machine directly} & \textbf{CUA + recompile under verify-gate} & \textbf{Verified corpus growth + dedup-replacement} \\
\bottomrule
\end{tabular}
\end{table}

\textbf{Computer-using agents.} A rapidly growing line of work builds agents that operate real software through its graphical interface---clicking, typing, and scrolling---rather than through bespoke APIs~\cite{anthropic2024computeruse}. Early web agents learned from demonstrations or reinforcement on simulated sites~\cite{nakano2021webgpt,yao2022webshop,gur2024webagent}, and benchmarks such as Mind2Web~\cite{deng2023mind2web}, WebArena~\cite{zhou2024webarena}, VisualWebArena~\cite{koh2024visualwebarena}, and WebVoyager~\cite{he2024webvoyager} pushed toward realistic, open-ended websites. A parallel thread grounds vision-language models directly on screen pixels and accessibility trees: SeeAct~\cite{zheng2024seeact}, CogAgent~\cite{hong2024cogagent}, SeeClick~\cite{cheng2024seeclick}, and UI-TARS~\cite{qin2025uitars} learn to localize and act on UI elements. The same paradigm now spans mobile~\cite{zhang2023appagent,wang2024mobileagent,rawles2023androidwild,rawles2024androidworld}, desktop and OS control~\cite{niu2024screenagent,wu2024oscopilot,zhang2024ufo,bonatti2024windows,xie2024osworld}, and software engineering~\cite{yang2024sweagent,jimenez2024swebench}, with reinforcement learning~\cite{bai2024digirl} and general-assistant benchmarks~\cite{mialon2024gaia} measuring progress. PreAct is agnostic to which such agent fills its CUA-fallback slot; we use AndroidWorld's T3A and Anthropic's Computer-Use API, but the harness wraps any of them.

\textbf{The reason--act loop and its cost.} Almost all of these agents follow the observe--reason--act paradigm popularized by ReAct~\cite{yao2023react}, layered on chain-of-thought and search-style reasoning~\cite{wei2022cot,yao2023tot,wang2023selfconsistency}, tool use~\cite{schick2023toolformer}, self-reflection~\cite{shinn2023reflexion}, and autonomous task loops~\cite{autogpt2023,sumers2024coala}. This loop is exactly what makes repeated execution expensive: every step pays for fresh perception and reasoning even when the task is one the agent has solved before~\cite{lu2024proactive}. PreAct attacks this cost directly by replacing the loop, on familiar tasks, with deterministic graph execution.

\textbf{Skill libraries and experiential memory.} A second response to repetition is to remember. Skill systems accumulate reusable behaviours---Voyager grows a library of code skills~\cite{wang2024voyager}, SAGE and Skill-Pro parameterize learned skills~\cite{liang2024sage,mi2026skillpro}, and ExpeL distills experience into rules~\cite{zhao2024expel}---but each skill still runs inside an LLM-bound loop at execution time. More broadly, agent-memory systems store and retrieve past interactions: generative-agent memory streams~\cite{park2023generative}, operating-system-style memory~\cite{packer2023memgpt}, and production memory stacks~\cite{chhikara2025mem0,xu2025amem} (surveyed in~\cite{zhang2024memorysurvey}), typically consulted by retrieval-augmented generation~\cite{lewis2020rag}. PreAct's corpus is a memory in this sense, but what it stores is \emph{executable}: recall and action are the same step. User-as-Code~\cite{li2026userascode} represents a user's accumulated state as typed Python objects and the rules over it as executable constraints, so representing the user and reasoning about the user share one medium. The two papers differ in what the code \emph{is}: there it is typed state plus constraint predicates queried over a user model; here it is a verified state-machine \emph{action program} whose execution is itself the recall. Scaling such reuse, rather than re-deriving behaviour, is consistent with the ``bitter lesson'' that general mechanisms outrun hand-engineered ones~\cite{sutton2019bitter}.

\textbf{Code as the agent's action and representation.} A third thread expresses agent behaviour as code. PAL, Program-of-Thoughts, and CoCoGen offload reasoning to executable programs~\cite{gao2023pal,chen2023pot,madaan2022cocogen}; Code-as-Policies and CodeAct make code the agent's action space~\cite{liang2023codeaspolicies,wang2024codeact}; Chain-of-Code emulates execution~\cite{li2024chainofcode}; and TroVE and LATM induce reusable, verifiable toolboxes from solved tasks~\cite{wang2024trove,cai2023latm}. PreAct shares the conviction that code is the right substrate for reliable reuse, but its compiled artifact is a \emph{verified state machine executed directly}, not a flat script regenerated and re-interpreted by an LLM each time.

\textbf{Internalized vs.\ externalized world models.} A fourth thread learns the environment's \emph{dynamics}---a model that predicts the next observation, or a tool's response, from the current state and a candidate action. Classical world models learn these dynamics in network parameters and plan inside the learned simulator~\cite{ha2018worldmodels,hafner2023dreamerv3}; the idea has since migrated to language agents that use the LLM itself as a world model to predict action outcomes for planning~\cite{hao2023rap}, including for web navigation, where an agent learns environment dynamics to anticipate the effect of a click before committing to it~\cite{chae2024wma}. PreAct can be read as a world model of a different kind. Each stored program is an \emph{externalized, verifiable} model of one task's dynamics---its state predicates are checkable claims about what the screen will be, and its transitions are the actions that move between those states---rather than dynamics baked into weights. The difference is where the learning lives and whether it can be trusted: an internalized world model predicts transitions implicitly and can hallucinate them with no way to check against the live environment, whereas PreAct's verify-before-store gate (\S\ref{sec:arch-gate}) is exactly a test that the learned dynamics still hold before a program is trusted, and a program can be inspected, replaced, or repaired without retraining. The price is generality: these are task-specific models grounded in concrete selectors, not one dense dynamics model, and they do not transfer across task families (\S\ref{sec:eval-ood}).

\textbf{Compiling and replaying workflows.} Closest to PreAct are systems that compile or cache concrete workflows. Robotic process automation has long replayed recorded UI scripts deterministically~\cite{aalst2018rpa}; Compiled-AI and ActionEngine compile agent behaviour into deterministic code, the latter generating flat Python from a state machine built by untargeted crawling~\cite{compiledai2026,actionengine2026}; and concurrent record-and-replay systems---Muscle-Mem~\cite{musclemem}, AgentRR~\cite{agentrr2025}, and Workflow-Use~\cite{workflowuse}---cache a linear action sequence and fall back to the agent on a cache miss. We compare against these directly. The gap PreAct closes is the same one at two levels: the stored artifact is never checked against the live UI---not per step as it replays (so a blind action sequence walks straight into a changed screen), and not before it is trusted as a stored behaviour (so the artifact is appended, never refined or revalidated). PreAct adds verification at both points. Table~\ref{tab:related} summarizes the comparison.

\textbf{Running the state machine directly.} What distinguishes PreAct is the second column (it runs the state machine directly) combined with the fourth (recompile under the verify-gate). ActionEngine builds a state machine but emits flat Python from it; once the script is generated the state machine is thrown away, and what actually runs is that script. PreAct instead executes the state machine itself: each state's verification predicate is evaluated against the live UI, and each transition is dispatched to the underlying action backend (XPath click, JSON action, etc.). This direct execution enables (a) per-state verification and graceful fallback, (b) in-place patching when a transition fails, and (c) the verify-before-store gate (which we will argue is empirically necessary for monotonicity).

\textbf{What grows with experience.} The deeper distinction concerns \emph{what grows} as the agent gains experience. In Muscle-Mem, AgentRR, and Workflow-Use, the cache or experience store is append-only: every captured trajectory adds a new entry without affecting the existing ones. In ActionEngine, the state machine is built once via crawling and the flat Python is regenerated wholesale on each rebuild; there is no notion of incremental refinement. PreAct's compile-extend-replace loop, by contrast, allows the corpus to \emph{mutate}: a fresh compile with the same dedup signature replaces an older program if it passes the verify-gate. The corpus does not just accumulate---it refines its representation of repeated tasks as the agent encounters them across different parameter values, environments, and sessions. This is the property we mean by ``self-extending executable code corpus'' (\S\ref{sec:intro}): not an append-only cache, but a mutable, verified, directly-executable code library.

\section{System Architecture}
\label{sec:arch}

PreAct rests on a single discipline applied at two timescales: \emph{predict, then check against the live screen before trusting the prediction}. At run time the replayer checks each state's predicate before it fires the transition out of that state---observe first, then act---which is what makes direct, model-free execution trustworthy (\S\ref{sec:arch-loop}). At store time the same check is run once more, over the whole freshly compiled program, before it is allowed into the corpus---which is what keeps the growing corpus honest (\S\ref{sec:arch-gate}). The per-step check is the more primitive of the two: the store-time gate is simply that same predicate-checking replay, run from a clean state and confirmed by an independent evaluator. We describe the stored artifact (\S\ref{sec:arch-program}), the loop that runs and extends it (\S\ref{sec:arch-loop}), and the two points at which it verifies.

\subsection{What the agent stores: a state-machine program}
\label{sec:arch-program}

\begin{figure}[t]
\centering
\includegraphics[width=0.96\linewidth]{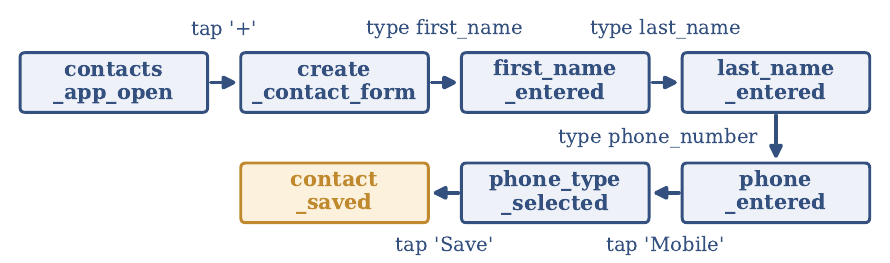}
\caption{The compiled ``add a contact'' program of Listing~\ref{lst:contact} drawn as a state machine (read top row left-to-right, then bottom row right-to-left). Each box is a state---its verification predicate, which must match the live screen, is given in Listing~\ref{lst:contact}---and each arrow is a transition holding an action. The harness executes this graph directly; it is not regenerated into a flat script.}
\label{fig:program-graph}
\end{figure}

We start from the artifact, because in PreAct the agent runs that artifact directly rather than regenerating a script from it. Listing~\ref{lst:contact} is an \emph{actual} program from the corpus, compiled while the agent ran the \texttt{ContactsAddContact} task from the AndroidWorld benchmark~\cite{rawles2024androidworld}---the trace was the instance ``Create a new contact for Emilia Gonzalez.'' It is a small graph, drawn in Figure~\ref{fig:program-graph}: each \emph{state} carries a verification predicate (an accessibility-tree pattern that must hold on the live screen for the agent to believe it is in that state) and each \emph{transition} carries an action. The concrete values typed during the trace are lifted to \emph{parameters}---here \texttt{first\_name}, \texttt{last\_name}, \texttt{phone\_number}, listed in the metadata---so this single program serves every future ``add a contact'' request by binding the parameters to each new contact's values.

\begin{lstlisting}[caption={The complete compiled program for ``add a contact,'' taken verbatim from the corpus (the AndroidWorld \texttt{ContactsAddContact} task). Only the schema-default \texttt{null} fields of each action object are omitted; everything else---state ids, verification predicates with their timeouts, per-state descriptions, and the full resource-id selectors---is reproduced exactly. Figure~\ref{fig:program-graph} draws the same program.},label={lst:contact},captionpos=b,basicstyle=\ttfamily\footnotesize,language={}]
{
  "metadata": {
    "program_id": "ab4390a9-1902-47a1-821b-ebddc7f4010a",
    "task_description": "Create a new contact for Emilia Gonzalez. Their number is +14240925675.",
    "application_context": "com.google.android.contacts",
    "parameters": ["first_name", "last_name", "phone_number"]
  },
  "states": [
    { "id": "contacts_app_open",
      "verification": {"type": "expect_element", "xpath": "resource_id=com.google.android.contacts:id/floating_action_button", "timeout_ms": 5000},
      "description": "Contacts app is open showing the main list with the Create-Contact FAB visible" },
    { "id": "create_contact_form",
      "verification": {"type": "expect_element", "xpath": "class=android.widget.EditText&&hint=First name", "timeout_ms": 3000},
      "description": "New-contact creation form is open with the First name field visible" },
    { "id": "first_name_entered",
      "verification": {"type": "expect_element", "xpath": "class=android.widget.EditText&&hint=Last name", "timeout_ms": 2000},
      "description": "First name has been entered; Last name field is available" },
    { "id": "last_name_entered",
      "verification": {"type": "expect_element", "xpath": "class=android.widget.EditText&&hint=Phone", "timeout_ms": 2000},
      "description": "Last name has been entered; Phone field is available" },
    { "id": "phone_entered",
      "verification": {"type": "expect_element", "xpath": "resource_id=android:id/text1&&text=Mobile", "timeout_ms": 2000},
      "description": "Phone number has been entered; phone-type selector (Mobile) is visible" },
    { "id": "phone_type_selected",
      "verification": {"type": "expect_element", "xpath": "resource_id=com.google.android.contacts:id/toolbar_button&&text=Save", "timeout_ms": 2000},
      "description": "Phone type Mobile selected; Save button is visible in the toolbar" },
    { "id": "contact_saved",
      "verification": {"type": "terminal_state", "timeout_ms": 5000},
      "description": "Contact has been saved successfully and the task is complete" }
  ],
  "transitions": [
    { "from_state": "contacts_app_open", "to_state": "create_contact_form",
      "action": {"type": "action_navigate", "text": "Contacts"} },
    { "from_state": "contacts_app_open", "to_state": "create_contact_form",
      "action": {"type": "action_click", "target": "resource_id=com.google.android.contacts:id/floating_action_button"} },
    { "from_state": "create_contact_form", "to_state": "first_name_entered",
      "action": {"type": "action_type", "target": "class=android.widget.EditText&&hint=First name", "parameter_name": "first_name"} },
    { "from_state": "first_name_entered", "to_state": "last_name_entered",
      "action": {"type": "action_type", "target": "class=android.widget.EditText&&hint=Last name", "parameter_name": "last_name"} },
    { "from_state": "last_name_entered", "to_state": "phone_entered",
      "action": {"type": "action_type", "target": "class=android.widget.EditText&&hint=Phone", "parameter_name": "phone_number"} },
    { "from_state": "phone_entered", "to_state": "phone_type_selected",
      "action": {"type": "action_click", "target": "resource_id=android:id/text1&&text=Mobile"} },
    { "from_state": "phone_type_selected", "to_state": "contact_saved",
      "action": {"type": "action_click", "target": "resource_id=com.google.android.contacts:id/toolbar_button&&text=Save"} }
  ]
}
\end{lstlisting}

The three values typed during the trace are recorded as the parameters \texttt{first\_name}, \texttt{last\_name}, \texttt{phone\_number} and re-bound to new values on every retrieval, so the one program generalizes across contacts. The two transitions out of \texttt{contacts\_app\_open} (an \texttt{action\_navigate} that opens Contacts and the \texttt{action\_click} on the create button) are a small redundancy the compiler emitted and the replayer tolerates; both reach \texttt{create\_contact\_form}.

Formally a program is a tuple $P=(S,T,M,V)$: states $S$ (id, description, verification predicate), transitions $T$ (each $(s_i, s_j, a)$ firing action $a$ to move from $s_i$ to $s_j$), metadata $M$ (task description, app context, parameter schema, and a \emph{dedup signature} that identifies which task family the program serves), and data-extraction predicates $V$ for question-answering tasks (e.g.\ \texttt{inspect\_text}). The full action vocabulary is in Appendix~A. This graph form buys three things a flat script cannot: \textbf{(1)} per-state verification, so the agent knows immediately when an action did not have the intended effect; \textbf{(2)} branching, since one state can have several outgoing transitions chosen by predicate (a delete-confirmation dialog that appears only on some app versions); and \textbf{(3)} in-place extension, since a new state and transition can be spliced in without regenerating the whole program.

\subsection{The loop: select, replay, fall back, verify}
\label{sec:arch-loop}

Figure~\ref{fig:arch} shows how the harness uses such programs. For a goal $T$:

\begin{enumerate}[leftmargin=2em]\tightlist
\item \textbf{Select.} The Program Selector queries the corpus with $T$ and returns either a candidate program $P$ or ``no candidate.''
\item \textbf{Replay.} If $P$ exists, the Replayer walks its graph---checking each state's predicate against the live screen and firing each transition's action. ``Add Emilia Gonzalez'' on Tuesday is solved here, in seconds, with no language-model call.
\item \textbf{Fall back.} If a predicate is false or an action errors, the harness hands control to the full agent (CUA), which continues from the live screen and records a fresh trace.
\item \textbf{Verify and store.} On success the trace is compiled into a new program $P'$; the verify-before-store gate (\S\ref{sec:arch-gate}) re-runs $P'$ from a clean state and stores it---replacing any program with the same dedup signature---only if it independently passes.
\end{enumerate}

The static harness code defines this loop; the corpus is what grows, and step~4's gate is what lets it grow \emph{monotonically} rather than accumulate programs that run but do not work. Algorithm~\ref{alg:harness} makes the loop precise.

\textbf{Replay verifies before it acts.} The distinction from blind record-and-replay lives in step~2, and it is the property we most want to make explicit. A classical RPA script or a cached action trace fires its stored actions in order, trusting that the screen is wherever it left off; PreAct instead checks each state's verification predicate against the live screen \emph{before} it performs the transition out of that state---observe first, then act. When the prediction holds---the expected element is present, the page is the one the program expects---the action fires immediately, with no language-model call, which is where the speed comes from. When it does not---an error dialog has appeared, an element is missing, the app has navigated somewhere unexpected---the predicate fails, replay halts instead of blindly clicking into a wrong state, and control passes to the CUA fallback (step~3). The agent finds a path from the live screen, and that path is compiled and, if it passes the gate, stored, so the program accrues new branches over time---success states, error-handling states, version-specific dialogs---covering more of the task's real state space each time it is exercised. This verify-before-act discipline, applied at \emph{every} step rather than only at the end, is what makes replay reliable enough to trust with no model in the loop; it is the difference between an agent that fluently re-executes a known task and one that blindly repeats a recording until it walks off a cliff. We isolate this run-time verification's own contribution---against a flat-script runtime that strips it---in \S\ref{sec:eval-not-loadbearing}.

\begin{figure}[t]
\centering
\includegraphics[width=0.92\linewidth]{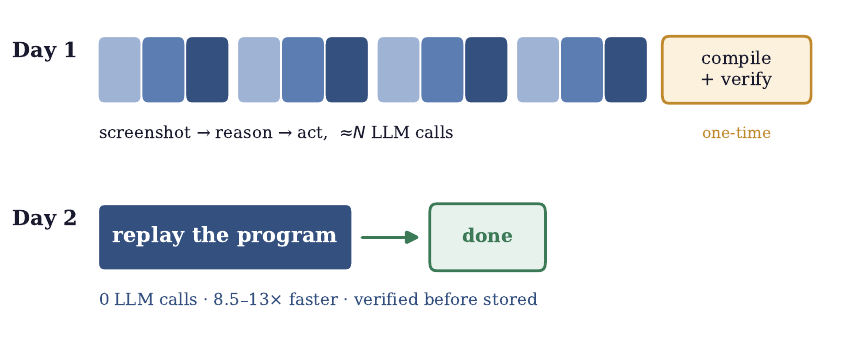}
\caption{Where the cost goes. The first time PreAct sees a task it runs the full agent (top: $\approx N$ rounds of observe--reason--act) and pays a one-time compile-and-verify premium. Every later run replays the stored program (bottom): no language-model calls, an order of magnitude faster, and already verified before it was stored.}
\label{fig:daytimeline}
\end{figure}

\begin{algorithm}[t]
\caption{PreAct verified compile-extend-replace loop}
\label{alg:harness}
\begin{algorithmic}[1]
\Require Task goal $T$, environment $\textit{env}$, program corpus $C$, replay-coverage threshold $\tau$ (default 0.5)
\State $P \gets \textsc{Selector}(T, C)$ \Comment{LLM-agentic; may return $\bot$}
\State $r \gets \textsc{None}$
\If{$P \neq \bot$}
  \State $r \gets \textsc{Replay}(P, \textit{env})$ \Comment{walk graph; verify each state predicate}
  \If{$r.\textit{success} \land r.\textit{cov} > \tau$}
    \State \textbf{return} $r$ \Comment{warm path: trusted replay}
  \EndIf
\EndIf
\State $r \gets \textsc{CUA}(T, \textit{env}, r)$ \Comment{hybrid: when $r{\neq}\textsc{None}$, CUA continues from the partial-replay state}
\If{$\neg r.\textit{success}$} \textbf{return} $r$ \EndIf
\State $P' \gets \textsc{Compile}(r.\textit{trace})$ \Comment{LLM-driven trace $\to$ state-machine}
\State $\textit{env}' \gets \textsc{Reset}(\textit{env}, T)$ \Comment{independent re-evaluation environment}
\State $r' \gets \textsc{Replay}(P', \textit{env}')$
\State $\textit{score}' \gets \textsc{Evaluate}(\textit{env}', T)$
\If{$r'.\textit{success} \land \textit{score}' \geq 1.0$} \Comment{double gate}
  \State $C \gets \textsc{Upsert}(C, P')$ \Comment{insert by dedup signature; replace if collision}
\EndIf
\State \textbf{return} $r$
\end{algorithmic}
\end{algorithm}

Three properties of Algorithm~\ref{alg:harness} are worth highlighting. First, \textbf{the harness's only side effect on $C$ is line 16's $\textsc{Upsert}$ call, gated by line 15's double predicate}: programs enter the corpus only after independent re-validation. Second, \textbf{the corpus grows monotonically in expectation}: each $\textsc{Upsert}$ call adds a program that has independently re-passed both replay and evaluation, so in expectation the corpus quality does not decrease across $\textsc{Upsert}$ events (a stored program can still fail on a later parameter binding---\S\ref{sec:eval-ood}---so this is an expectation, not a guarantee). Third, \textbf{the corpus mutates via $\textsc{Upsert}$, not just append}: a fresh program with the same dedup signature replaces an older one, allowing the corpus to refine its representation of repeated tasks rather than just accumulate variants.

\subsection{The verify-before-store gate}
\label{sec:arch-gate}

\begin{figure}[t]
\centering
\includegraphics[width=0.95\linewidth]{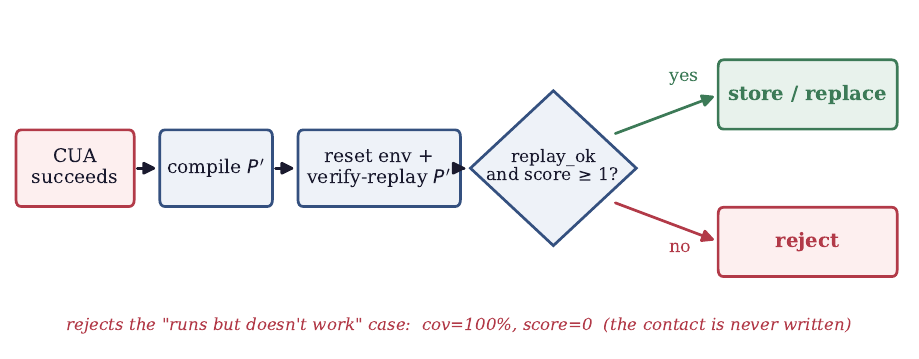}
\caption{The verify-before-store gate. A freshly compiled program is re-run from a clean state and re-scored by an independent evaluator; it enters the corpus only if it both replays cleanly \emph{and} passes. This is what rejects the ``runs but doesn't work'' programs---the ones that replay to full coverage yet leave the task unsolved.}
\label{fig:gate-flow}
\end{figure}

The gate is the run-time check of \S\ref{sec:arch-loop} applied once more---this time to the whole program, at the moment of storage---and it is what separates PreAct from a blind cache. When the agent succeeds and the trace is compiled into a candidate $P'$, the harness does \emph{not} trust it. It resets the environment to the task's clean initial state, replays $P'$ there (the same predicate-checking replay used on every warm run), and asks the benchmark's own evaluator whether the task is solved (Figure~\ref{fig:gate-flow}). $P'$ is stored only if \emph{both} hold: the replay ran to its terminal state without error, \emph{and} the evaluator scores it a pass.

Both halves earn their place. Checking the evaluator's pass alone lets through programs whose actions silently no-op (a malformed action the backend swallows) while the environment happens to still hold passing state from an earlier step---so we additionally require the replay to report its own success. The converse and more important failure mode is a clean replay that nonetheless does not pass: the program executes every action to completion (100\% coverage) yet the end state does not satisfy the evaluator---the contact form was filled and ``Save'' was tapped, but a stale field meant no contact was written. We call this the \textbf{cov$=$100\%/score$=$0 lossy replay}, and \S\ref{sec:eval-gate} shows it is exactly what the gate must catch.

One assumption is built into this design: because the gate verifies by \emph{re-running} the program rather than inspecting it, the verify step performs the task's actions a second time. This is safe in our benchmarks, where every task (and every verify-replay) starts from a scripted environment reset, but it presumes a resettable or otherwise idempotent task; tasks with irreversible side effects (sending a message, charging a card) would need a side-effect-free verification path instead. We return to this in limitation L6 (\S\ref{sec:limitations}).

\subsection{Selector and fallback}
\label{sec:arch-selector}

Two supporting components complete the loop. The \textbf{Program Selector} decides which stored program (if any) applies to a new goal; by default it is an LLM that reads the goal and each program's description and parameter schema and chooses via a tool call, but \S\ref{sec:eval-selector-ablation} shows a plain embedding retriever does just as well---the choice of selector makes little difference to the result. The \textbf{CUA fallback} is the full agent, invoked whenever the selector finds no candidate or a replay fails; we use AndroidWorld's T3A agent on mobile and Anthropic's Computer-Use API on desktop and web. The production fallback runs the \emph{verbatim} upstream agent prompt with no PreAct-specific additions (\S\ref{sec:eval-not-loadbearing})---the harness, not the prompt, is what makes the system work.

\section{Empirical Validation}
\label{sec:eval}

\subsection{Setup}
\label{sec:eval-setup}

\textbf{Benchmarks.} We evaluate on (a) AndroidWorld~\cite{rawles2024androidworld} ``official-15'' subset (15 tasks across 8 application domains; the curated subset used by the upstream T3A baseline), (b) OSWorld~\cite{xie2024osworld} ``test\_tiny'' subset (6 tasks across Chrome, LibreOffice Calc, and LibreOffice Writer), and (c) a 12-task shopping\_admin subset of WebArena~\cite{zhou2024webarena} (string-match answer-extraction tasks on a Magento e-commerce admin panel).

\textbf{Models.} For Android CUA, we use Gemini 3 Flash via the multi-provider port; this is roughly 10$\times$ cheaper than Claude Sonnet 4.6 at equivalent SR (we confirm this with cross-model replication in \S\ref{sec:eval-monotonic}). For OSWorld and WebArena CUA, we use Claude Sonnet 4.6 via Anthropic's Computer-Use API. For program compilation and selector reasoning, we use Claude Sonnet 4.6 throughout.

\textbf{Container.} AndroidWorld runs in the public \texttt{android\_world} container image, extended with a small patch that exposes the device's screenshot and accessibility tree to the agent (Appendix~C). OSWorld uses the standard \texttt{osworld-docker} image via its environment provider. WebArena uses the canonical \texttt{shopping\_admin} Magento Docker stack.

\textbf{Multi-seed protocol.} For each seed in \{42, 100, 1337, 2024, 7777\}, we set the existing corpus aside (preserving prior state) and start from a fresh empty corpus. We then run cold (empty corpus) followed by warm (cold-built corpus) on the same task list.

\textbf{Cost and protocol.} The full validation push ran in $\approx 50$ hours of autonomous container time for $\approx\$30$--$35$ in LLM spend. Every claim below uses paired cold/warm runs: ``cold'' starts from an empty corpus; ``warm'' reuses the corpus the cold run built on the same tasks. This is the setting the rerun crisis lives in---the same user, the same tasks, a second time.

\textbf{Metrics.} We report task success rate \emph{and} the cost of achieving it---wall-clock time, language-model token spend, and replay coverage---rather than success alone. The three benchmarks we build on score only whether a task passes, which makes an agent that re-derives every repeated task indistinguishable from one that has learned to replay it. Because PreAct's whole premise is that repeated work should get cheaper, not just stay correct (\S\ref{sec:intro}), we treat this efficiency dimension as a first-class outcome: warm-run speedup and the one-time compile-and-verify overhead are measured quantities (Figures~\ref{fig:rerun}, \ref{fig:gatecost}), not illustrative ones.

A one-paragraph map of what follows. Two harness mechanisms move the success rate: the \emph{verify-before-store gate} (\S\ref{sec:eval-gate}) and the \emph{cache-miss fallback} (\S\ref{sec:eval-baselines}). The runtime add-ons we tried---prompt content, runtime guardrails, and the choice of selector---do not (\S\ref{sec:eval-not-loadbearing}); the one remaining knob, the step-budget cap, trades success rate for wall-clock time by design rather than adding any. \textbf{The harness is what works.}

\subsection{Repeated runs get faster}
\label{sec:eval-monotonic}

\begin{figure}[t]
\centering
\includegraphics[width=0.60\linewidth]{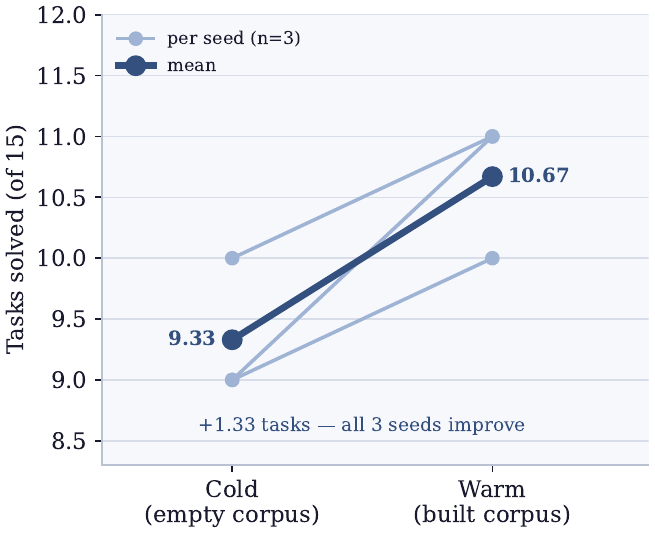}
\caption{\textbf{Cold$\to$warm refinement is monotonic.} AndroidWorld official-15 success rate, three Gemini 3 Flash seeds with the corpus reset per seed. Every seed improves (mean $9.33 \to 10.67$ of 15, $\Delta = +1.33$); none regresses. Per-task detail in Appendix~B.}
\label{fig:monotonic}
\end{figure}

The basic promise is that the second time the agent sees a task, it is faster and no worse. Figure~\ref{fig:monotonic} shows the success rate rising from cold to warm on AndroidWorld official-15 across three Gemini 3 Flash seeds (corpus reset per seed): every seed improves, mean $+1.33$ tasks, with \emph{no} seed regressing. The gain is not free retrieval of luck---on the warm runs the agent handles a majority of tasks by replaying a stored program rather than re-deriving it. Of the 13 tasks that clear setup on seed~42, 8 shift from fresh-agent solving on cold to replay-or-hybrid on warm (versus all 13 re-derived by the fresh agent on cold), which is the corpus doing its job.

\begin{figure}[t]
\centering
\includegraphics[width=0.58\linewidth]{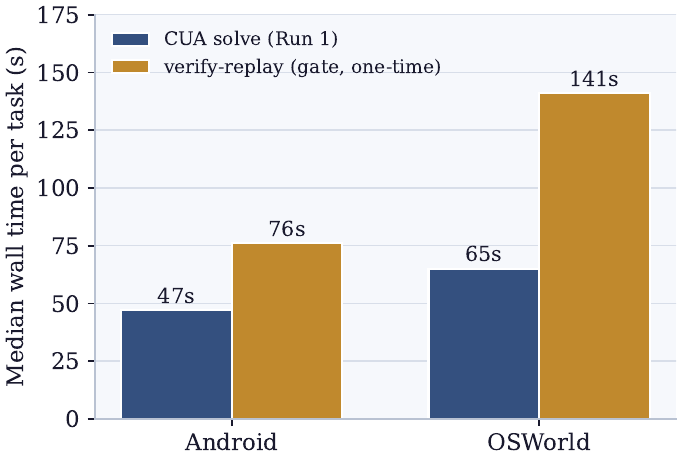}
\caption{The gate's one-time cost. Verifying a freshly compiled program (re-running it from a clean state) costs about as much as---or, on both platforms measured, somewhat more than---the original solve ($+162\%$ wall time on Android, $+217\%$ on OSWorld), the gap being largest on OSWorld where both the solve and the verify-replay are slowest. This is paid once per stored program and amortized over every warm run.}
\label{fig:gatecost}
\end{figure}

\textbf{The speedup is real.} A replayed program carries no per-step language-model call, so warm runs are both cheaper and faster: a served replay makes essentially no model calls (its LLM cost collapses to near zero), and on WebArena completes $8.5$--$13\times$ faster in wall-clock time than the corresponding fresh-agent solve (Figure~\ref{fig:rerun}), while on Android a warm task that hits the corpus finishes in seconds. The one-time price is the compile-and-verify step, which costs roughly as much as one more solve ($+162\%$ wall time on Android, $+217\%$ on OSWorld per stored program; Figure~\ref{fig:gatecost}); it is repaid the first time the task recurs. Figure~\ref{fig:daytimeline} traces where this cost goes: a one-time premium on the first solve, then near-free replays on every later run.

\textbf{The result is not model-specific.} Swapping Gemini 3 Flash for Claude Sonnet 4.6 on the same subset gives the same $73.3\%$ (11/15) and---tellingly---the \emph{same} three stable failures (BrowserDraw, SystemBrightnessMax, SystemWifiTurnOn) across both models and all three seeds. These three are properties of the benchmark's UI (a canvas not exposed in the accessibility tree; a brightness slider that rejects scroll; a stale status-bar Wi-Fi icon), not of the agent's reasoning---they bound the benchmark, not the method (\S\ref{sec:disc-harness-det}).

\subsection{The verify-gate keeps the speedup from rotting}
\label{sec:eval-gate}
\label{sec:eval-webarena}

Reuse cuts both ways. The same corpus that makes warm runs cheap can make them \emph{wrong}, if it stores a program that runs but does not work. Recall the lossy contact program from \S\ref{sec:arch-gate}: it taps through the form and presses Save, replays at 100\% coverage, reports success---and yet writes no contact, because a stale field left the name blank. Store that program, and every future ``add a contact'' silently fails. The verify-before-store gate exists to keep such programs out. To show it is doing real work, we turn it off.

\begin{figure}[t]
\centering
\includegraphics[width=0.82\linewidth]{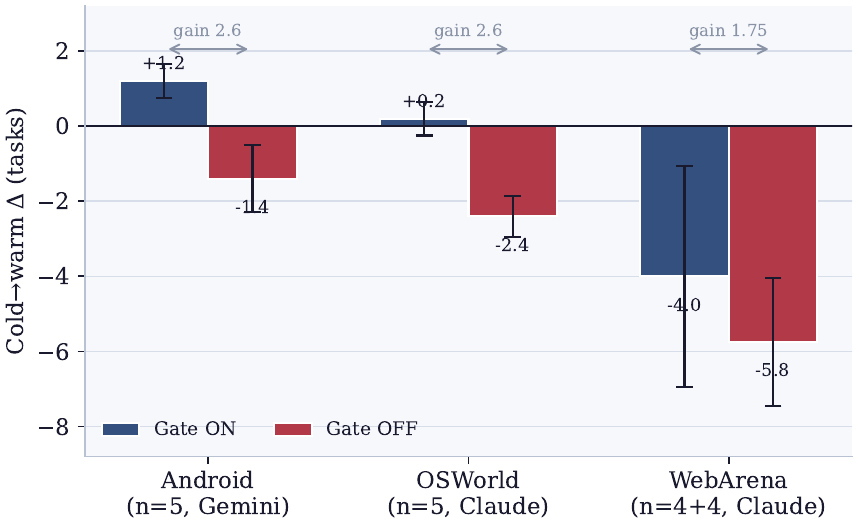}
\caption{\textbf{The gate is what makes repeated runs improve.} Mean cold$\to$warm success-rate change with the gate on (blue) versus off (red), $\pm 1$ s.d. On all three platforms the gate-on bar is higher: the gate makes the corpus improve the agent on mobile and desktop, and degrade it far less on the web (where most programs cannot be re-verified, \S\ref{sec:eval-webarena}); without it, the corpus \emph{degrades} the agent everywhere as lossy programs accumulate. The gap between the bars (the gate's marginal value) is $1.75$--$2.6$ tasks and points the same way everywhere---this is a property of the harness, not of any one platform. Per-seed and per-rep data in Appendix~B.}
\label{fig:diff}
\end{figure}

We ran the same $2{\times}2$ ablation---cold/warm $\times$ gate on/off---on all three platforms: AndroidWorld official-15 ($n{=}5$ seeds, Gemini), OSWorld test\_tiny ($n{=}5$ reps, Claude), and a 12-task WebArena shopping\_admin subset ($n{=}4{+}4$ reps, Claude). Table~\ref{tab:gate} gives the per-platform cold$\to$warm change in each condition; Figure~\ref{fig:diff} plots it.

\begin{table}[t]
\centering
\caption{Verify-gate ablation across three platforms. Each cell is the mean cold$\to$warm change in tasks solved ($\pm 1$ s.d.). With the gate \emph{on}, repeated runs hold or improve on mobile and desktop and regress less on the web; with it \emph{off}, every platform regresses further as lossy programs accumulate. ``Gate's gain'' is the gap between the two---the cost of not verifying. The gate-on $\Delta$ is higher than gate-off on all three platforms, strictly so in 13 of the 14 paired runs (the 14th, WebArena rep~4, is a tie). Per-seed and per-rep tables are in Appendix~B.}
\label{tab:gate}
\small
\begin{tabular}{@{}lccc@{}}
\toprule
Platform (model) & $\Delta$ gate ON & $\Delta$ gate OFF & Gate's gain \\
\midrule
AndroidWorld official-15 (Gemini, $n{=}5$) & $+1.2 \pm 0.45$ & $-1.4 \pm 0.89$ & $2.6$ \\
OSWorld test\_tiny (Claude, $n{=}5$) & $+0.2 \pm 0.45$ & $-2.4 \pm 0.55$ & $2.6$ \\
WebArena shopping\_admin (Claude, $n{=}4{+}4$) & $-4.0 \pm 2.94$ & $-5.75 \pm 1.71$ & $1.75$ \\
\bottomrule
\end{tabular}
\end{table}

\textbf{The gate decides whether reuse helps or hurts.} Read the table one row at a time. On mobile and desktop the gate-on agent improves or holds across repeated runs ($\Delta \geq 0$), while the gate-off agent \emph{loses} ground---it does the same tasks a second time and solves fewer of them, because the corpus is now seeded with programs that replay but do not work. The effect points the same way on all three platforms and in every non-tied paired run (13 of 14; the 14th, WebArena rep~4, is a tie); per-platform sign tests give $p \approx 0.031$ on Android and OSWorld and $p = 0.125$ on WebArena's three non-tied pairs. Pooling the consistent direction across platforms gives $p \approx 10^{-4}$ under a no-effect null; because the within-platform repetitions are correlated rather than independent, we read that pooled figure as a directional lower bound, and rest the claim on the per-platform tests and the consistent direction. That the gate's gain lands in the same $1.75$--$2.6$ task band despite three platforms, three models, and three evaluator styles suggests it reflects how often the \emph{compiler} emits a lossy program---a property of compilation, not of any platform (\S\ref{sec:disc-structural}).

\textbf{Why WebArena is low even with the gate on.} On WebArena the gate-on agent still regresses ($-4.0$), because its tasks are answer-extraction (``what was the top-selling product in 2022?''): almost every compiled program reads a value off a page that has since changed, so the gate---correctly---rejects nearly all of them, leaving the corpus empty and nothing to replay. The gate is behaving exactly as designed (it never stores a program that fails re-verification); the missing piece is what to do when it empties the corpus, which is the subject of \S\ref{sec:eval-baselines}.

\textbf{The failure, observed directly.} The failure the gate intercepts is directly observable. Under gate-off we logged five distinct programs (Table~\ref{tab:smoking-gun}) that replay to 100\% coverage, report success---and score 0 on the evaluator. The first is our running example: a stored ``add a contact'' program that taps through the whole form and presses Save, yet leaves no contact behind. On WebArena the same pattern is near-total: all 48 gate-off warm replays score 0. These are precisely what the gate exists to reject before storage.

\begin{table}[t]
\centering
\caption{Five reproducible ``runs-but-doesn't-work'' programs logged under gate-off: each replays its full action sequence (100\% coverage) yet the evaluator scores it 0. The first is the running example. With the gate on, these are rejected at compile time and never enter the corpus.}
\label{tab:smoking-gun}
\small
\begin{tabular}{@{}lll@{}}
\toprule
Platform & Task & What goes wrong \\
\midrule
Android & ContactsAddContact & Replay completes; no contact written \\
Android & MarkorCreateFolder & Replay completes; folder not at expected path \\
OSWorld & Chrome history clean & Replay completes; browser state mismatches \\
OSWorld & LibreOffice Calc formula & Replay completes; cell formula wrong \\
OSWorld & LibreOffice Calc chart & Replay completes; chart properties mismatch \\
\bottomrule
\end{tabular}
\end{table}

\subsection{The second mechanism: a fallback that closes the gap to record-and-replay}
\label{sec:eval-baselines}

\begin{figure}[t]
\centering
\includegraphics[width=0.78\linewidth]{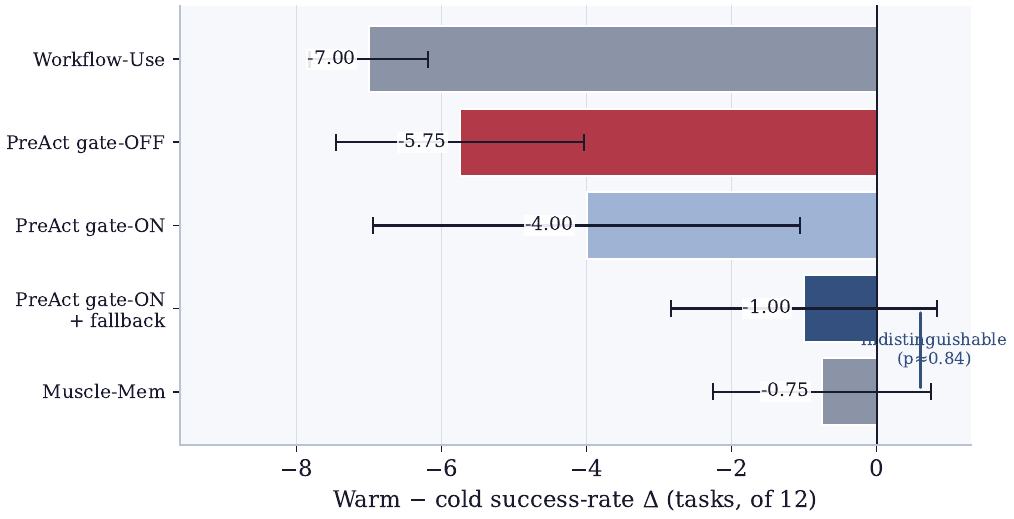}
\caption{\textbf{Closing the gap to Muscle-Mem on WebArena.} Warm$-$cold success-rate change ($\pm 1$ s.d.). As shipped, PreAct's gate alone ($-4.0$) trails Muscle-Mem ($-0.75$): when the gate empties the corpus, PreAct had nothing to fall back to, whereas Muscle-Mem's cache miss re-runs the full agent. Adding the same cache-miss fallback to PreAct ($-1.0$) makes the two statistically indistinguishable ($p \approx 0.84$). Workflow-Use and gate-off collapse to $0$ warm, for the same reason gate-off does in Figure~\ref{fig:diff}.}
\label{fig:baselines}
\end{figure}

The gate exposes a question: when it empties the corpus for a task (as it does on most WebArena tasks), what should the harness do? To answer it we compared PreAct against the strongest record-and-replay baseline, \textbf{Muscle-Mem}~\cite{musclemem}---which caches the linear action sequence, replays it blindly, and re-runs the full agent on any cache miss---and against \textbf{Workflow-Use}~\cite{workflowuse}, which stores per-task scripts with no verification. Same harness, tasks, and backend (Claude Sonnet 4.6), $n{=}4$ reps each (Figure~\ref{fig:baselines}, Table~\ref{tab:baselines}).

\begin{table}[t]
\centering
\caption{Head-to-head baselines on WebArena shopping\_admin 12-task subset, $n{=}4$ reps each. PreAct (gate-ON / gate-OFF) rows reproduced from \S\ref{sec:eval-webarena}; the bottom row is the cache-miss-to-CUA fallback harness fix (\S\ref{sec:eval-baselines-fallback}) that we propose and measure.}
\label{tab:baselines}
\footnotesize
\setlength{\tabcolsep}{4pt}
\begin{tabular}{@{}p{4.6cm}p{1.8cm}p{1.8cm}p{1.9cm}p{4.0cm}@{}}
\toprule
System & Cold SR & Warm SR & $\Delta$ (warm$-$cold) & Notes \\
\midrule
Muscle-Mem (blind linear cache) & $7.0 \pm 1.41$ & $6.25 \pm 0.96$ & $-0.75 \pm 1.50$ & Cache miss $\to$ CUA fallback per task \\
Workflow-Use (per-task scripts) & $7.0 \pm 0.82$ & $0 \pm 0$ & $-7.0 \pm 0.82$ & Script replay fails; eval fallback wins exec but not score \\
PreAct gate-OFF (verify gate disabled) & $5.75 \pm 1.71$ & $0 \pm 0$ & $-5.75 \pm 1.71$ & Stores every compile; warm SR collapses on lossy programs \\
PreAct gate-ON (verify gate enabled) & $6.0 \pm 0.82$ & $2.0 \pm 2.31$ & $-4.0 \pm 2.94$ & Gate rejects $\approx 83\%$ of compiles; bimodal warm SR (4/4/0/0) \\
\textbf{PreAct gate-ON + cache-miss CUA fallback} & $\mathbf{7.25 \pm 1.50}$ & $\mathbf{6.25 \pm 0.96}$ & $\mathbf{-1.0 \pm 1.83}$ & On gate-reject, run a fresh CUA on Run~2 (matches Muscle-Mem semantics) \\
\bottomrule
\end{tabular}
\end{table}

The honest reading of Table~\ref{tab:baselines} unfolds in two stages. \emph{Stage 1 (PreAct as-is vs.\ baselines).} At matched $n{=}4$, the as-shipped \textbf{PreAct gate-ON harness underperforms Muscle-Mem on this benchmark}: Muscle-Mem's warm-SR mean ($6.25/12 = 52\%$) is more than $3\times$ PreAct gate-ON's ($2.0/12 = 17\%$). The warm-$\Delta$ gap is $-0.75$ vs.\ $-4.0$, i.e.\ Muscle-Mem loses about $3.25$ fewer tasks than PreAct on the cold-to-warm transition. Workflow-Use, by contrast, behaves like PreAct gate-OFF (warm SR $0/12$ in every rep)---it stores compiled scripts without verification and, like gate-OFF, suffers complete warm-time collapse when the scripts encounter live-page selectors that don't match.

\textbf{Mechanism diagnosis: cache miss vs.\ verify-reject.} WebArena's two-run protocol replays the \emph{same task} on Run~2, so the relevant question is what happens when the stored artifact does not match the live page. Muscle-Mem's blind-replay attempt fails fast (typical $\sim 30$\,s before timing out on a missing selector), and the harness then falls through to a fresh CUA run that re-solves the task with the original LLM cost. PreAct's verify-before-store gate, by contrast, runs an independent verify-replay between Run~1 and Run~2; on these WebArena tasks the compiled state-machine predicates are brittle enough (e.g.\ \texttt{state=reviews\_index\_page} matched via XPath patterns) that they fail re-verification even on the cold-pass task, so the gate deletes the program and Run~2 has \emph{no artifact} to fall back to. \emph{Muscle-Mem's advantage on WebArena is not that its replay is more reliable---it isn't, with reps showing cache miss rates of $50$--$67\%$---but that its cache-miss path falls through cleanly to CUA, whereas PreAct's gate-reject path leaves Run~2 with no artifact at all.}

\subsubsection{Closing the Gap: Cache-Miss-to-CUA Fallback}
\label{sec:eval-baselines-fallback}

The diagnosis suggests a small change: when the gate rejects all compiles for a task, run a fresh CUA on Run~2 instead of returning ``no artifact''. We add a fallback branch that triggers a fresh CUA exploration whenever the verify-gate has emptied the corpus for the current task; we additionally tighten the gate's cleanup to delete every program added during the task's compile-and-verify cycle (Run-1 compile \emph{and} any programs the verify-replay's own CUA recovery happens to add), so a stale verify-recovery program cannot suppress the cache-miss branch. With this change, Run~2 fires a fresh CUA exactly when Muscle-Mem would: the bottom row of Table~\ref{tab:baselines}.

The result: \textbf{warm SR rises from $2.0/12$ to $6.25/12$, exactly matching Muscle-Mem's $6.25/12$, with the warm-$\Delta$ gap closing from $-4.0$ to $-1.0$.} Across the four reps the warm-SR sequence is $6, 7, 5, 7$ (vs.\ Muscle-Mem $5, 6, 7, 7$) and the per-rep warm-$\Delta$ values are $0, +1, -3, -2$ (vs.\ Muscle-Mem $-2, 0, +1, -2$). Welch's two-sided $t$-test on the warm-$\Delta$ distributions gives $|t| \approx 0.21$, $p \approx 0.84$: under any conventional threshold we cannot reject equality of means, i.e.\ PreAct-with-fallback and Muscle-Mem are statistically indistinguishable on this benchmark. The cold-SR comparison is similar ($7.25$ vs.\ $7.0$, $p > 0.5$).

This is a small-but-real \emph{positive} finding for the architecture: PreAct's gate-rejection signal is high quality (the rejections in the gate-ON-no-fallback condition were correct---the underlying compiled programs really were lossy), and the only thing missing was a fallback path. Critically, the fallback does \emph{not} compromise the gate's Android/OSWorld value: the gate continues to reject lossy compiles \emph{before} they enter the corpus, preserving monotonicity (\S\ref{sec:eval-gate}); the fallback only changes what happens on the cache-miss \emph{branch} of Run~2 on the platform where the gate's rejection rate is highest. The fallback also does not save tokens compared to Muscle-Mem on this benchmark---both pay the full CUA cost when the cache misses, so warm tokens are similar to cold ($\sim 65$\,k for fallback vs.\ $\sim 50$\,k cold; the slight increase comes from CUA re-exploring with a fresh browser session)---but it recovers the warm-SR property that the as-shipped harness was leaving on the table.

\textbf{Implication.} Across the three platforms tested, the harness needed two cooperating mechanisms: the verify-before-store gate to keep the corpus monotonic (\S\ref{sec:eval-gate}; Android $+2.6$, OSWorld $+2.6$, WebArena $+1.75$ task diff-of-deltas under the gate), \emph{and} a cache-miss-to-CUA fallback when the gate empties the corpus on a particular task (this section; WebArena $\Delta = -1.0$ matching Muscle-Mem's $-0.75$). The two-stage finding---``gate alone'' loses to Muscle-Mem on WebArena, but ``gate + cache-miss fallback'' matches Muscle-Mem on WebArena \emph{and} retains the gate's Android/OSWorld monotonicity---is the main architectural lesson of this section. PreAct enables this cache-miss fallback by default; disabling it recovers the gate-ON, no-fallback behavior reported as the $\Delta = -4.0$ measurement.

\subsection{Compile-LLM Robustness: Gemini vs Claude Compile}
\label{sec:eval-compile-llm}

\begin{figure}[t]
\centering
\includegraphics[width=0.72\linewidth]{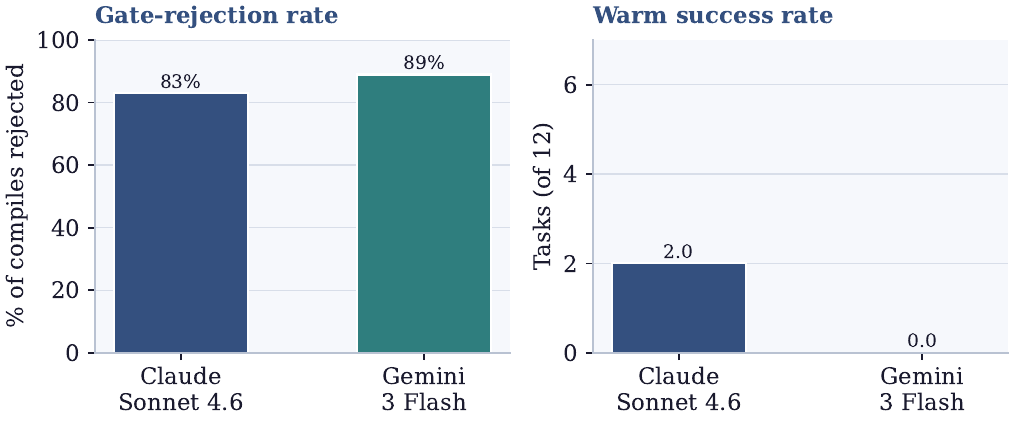}
\caption{The gate is mechanism-driven, not Claude-specific. Swapping the compile-step model from Claude to Gemini 3 Flash leaves the gate-rejection rate in the same band ($83\%$ vs.\ $89\%$); warm success rate depends only on what survives the gate.}
\label{fig:compile-llm}
\end{figure}

To probe whether the verify-gate's compile-rejection rate is Claude-specific or a property of the compile step itself, we swap the compile-step LLM from Claude Sonnet 4.6 to Gemini 3 Flash, leaving the CUA loop and selector on Claude. We ran $n{=}3$ WebArena reps on the same 12-task shopping\_admin subset (gate-ON, no cache-miss fallback, for clean comparison with the Claude-compile baseline in \S\ref{sec:eval-webarena}); Table~\ref{tab:compile-llm} and Figure~\ref{fig:compile-llm} report the result.

\begin{table}[t]
\centering
\caption{Compile-LLM robustness: gate behavior under Claude vs Gemini compile, WebArena 12-task shopping\_admin gate-ON (no cache-miss fallback).}
\label{tab:compile-llm}
\small
\begin{tabular}{@{}lcccc@{}}
\toprule
Compile LLM & Cold SR & Warm SR & $\Delta$ & Gate-rejection rate \\
\midrule
Claude Sonnet 4.6 ($n{=}4$ reps) & $6.0/12 \pm 0.82$ & $2.0/12 \pm 2.31$ & $-4.0 \pm 2.94$ & $\approx 83\%$ \\
Gemini 3 Flash ($n{=}3$ reps) & $6.33/12 \pm 0.58$ & $0/12 \pm 0$ & $-6.33 \pm 0.58$ & $\approx 89\%$ \\
\bottomrule
\end{tabular}
\end{table}

Cold SR is statistically indistinguishable across compilers ($6.0$ vs.\ $6.33$, $p > 0.5$), as expected since both use the same Claude Sonnet 4.6 CUA loop on Run~1. The gate's rejection rate is in the same band ($83\%$ with Claude vs.\ $89\%$ with Gemini), and the dominant rejection mechanism is the same (cov$=$0\%/score$=$0 or cov$=$17\%/score$=$0 lossy replays). Qualitatively, Gemini compiles tend to fail at very-first-step replay (cov$=$0\% modal) whereas Claude compiles run to completion but produce wrong answers (cov$=$100\% modal); both failure modes are caught by the gate. The Gemini-compile warm SR is $0/12$ on all three reps with $\sigma{=}0$---tighter than the Claude bimodal distribution ($4, 4, 0, 0$)---reflecting Gemini's higher rate of cov$=$0\% compiles that the gate rejects with $100\%$ consistency. \textbf{The verify-gate is mechanism-driven, not Claude-specific}: substituting Gemini 3 Flash at the compile step does not change the qualitative behavior (gate fires; lossy compiles get rejected; warm SR depends on what survives), and the quantitative rejection rate shifts only modestly ($83 \to 89\%$). Cross-model replication closes the compile-step concern noted under L1 in Appendix~E.

\subsection{What Did \emph{Not} Change the Result}
\label{sec:eval-not-loadbearing}

The AB experiments that came back \emph{negative} or marginal are just as useful: they rule out the obvious alternative explanations for why PreAct works. Detailed per-seed tables are in Appendix~F.

\begin{figure}[t]
\centering
\includegraphics[width=0.86\linewidth]{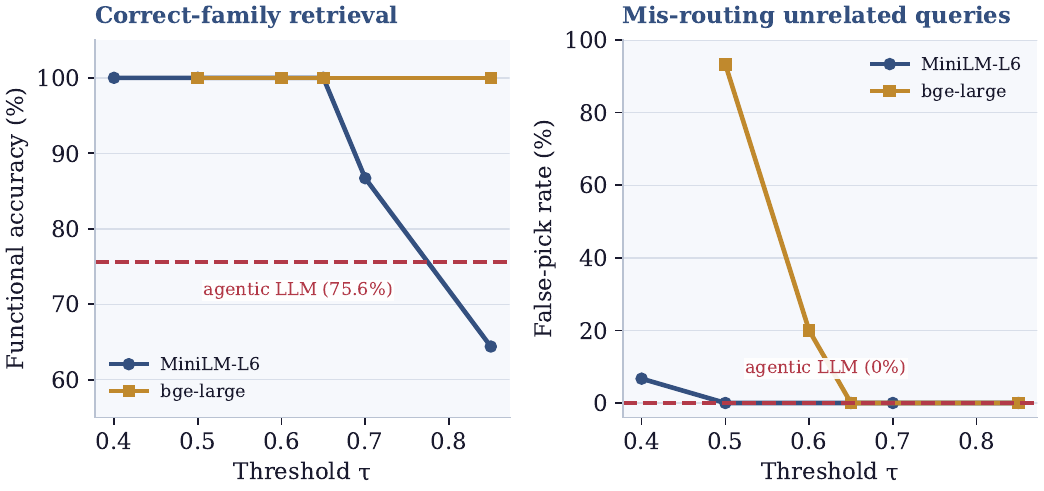}
\caption{\textbf{A plain embedding retriever matches the LLM-based selector.} On the 58-program corpus, both embedding backbones reach $100\%$ correct-family retrieval (left) at $0\%$ mis-routing of unrelated queries (right) once the similarity threshold $\tau$ is tuned, against the agentic selector's $75.6\%$ (dashed). The choice of selector thus barely affects retrieval quality---a tuned embedding retriever even edges out the agentic one, and the agentic selector's own misses are conservative no-picks that fall back to CUA rather than wrong retrievals; we keep it as the default only for its interpretable logs and tuning-free operation. Full sweep in Appendix~F.}
\label{fig:selector}
\end{figure}

\textbf{The choice of program selector barely matters.} We had assumed an LLM-based selector was necessary---deciding whether a stored program applies to a new task looked like a reasoning problem. It is not: a plain embedding retriever (MiniLM-L6 or bge-large) matches or beats the agentic selector on the 58-program corpus, reaching $100\%$ correct-family retrieval at $0\%$ mis-routing once its threshold is tuned, versus the agentic selector's $75.6\%$ (Figure~\ref{fig:selector}). We keep the agentic selector as the default only for its interpretable reasoning logs and because it needs no threshold tuning---not because it retrieves better.

\textbf{Prompt-level PreAct content is unused in production.} Our implementation includes three optional PreAct-specific guideline bullets, but the production CUA path never passes them to the agent. The 73.3\% Android SR is achieved with the verbatim upstream T3A prompt and zero PreAct prompt-level additions. The contribution on Android is the harness wrapping T3A---not the prompt.

\textbf{Code-level runtime guardrails are aggregate-neutral at $n{=}5$.} A $2{\times}2$ ablation of the guardrails (a double-tap before text entry, an image-task no-op cap, and scroll-exhaustion notes) gives \emph{identical} cold means ($10.2 = 10.2$) and warm means $11.0$ (on) vs.\ $10.6$ (off), a $0.4$-task gap well within seed-level variance (Figure~\ref{fig:guardrails}; Table~\ref{tab:guardrails}). Per-task analysis shows the double-tap helps the audio-record-with-filename task (a populated filename dialog) but the gain is cancelled by minor timing penalties on Camera tasks.

\textbf{Run-time verification (state-machine vs.\ flat-script): real but small on these benchmarks.} This ablation isolates the run-time half of PreAct's verification principle (\S\ref{sec:arch-loop}): it strips the per-state predicate checks, leaving a flat script that fires its actions blindly the way a recorded macro would. State-machine vs.\ flat-script (Figure~\ref{fig:arch-ab} and Table~\ref{tab:arch-ab} in Appendix~F) gives flat-script $10.6 \pm 1.14$ ($n{=}5$ seeds) vs.\ the verified runtime $11.67 \pm 0.58$ on the 3 matched seeds; matched-seed $\Delta = +0.67$ tasks favoring verification, all 3 paired comparisons in the non-regressive direction (sign-test $p = 0.125$, suggestive). The wins concentrate on Camera tasks, where verification catches mismatches that linear execution walks past. The effect is small here for a telling reason: per-step verification pays off exactly when the screen deviates from the program's expectation, and these short-horizon benchmark tasks rarely throw an anomaly mid-replay---so the mechanism shows its value more in \emph{how} a run fails (a clean halt-and-fall-back rather than clicking on into a wrong state) than in raw success rate. We therefore treat run-time verification as the enabling design choice that makes trustworthy direct replay possible, not as a large SR lever; the strongly-validated lever is the store-time gate (1.75--2.6-task diff-of-deltas).

\textbf{Dynamic step-budget cap: a deliberate speed/accuracy tradeoff.} Unlike the add-ons above, the step-budget cap is \emph{not} neutral---it is the one knob in this section that moves the success rate, and it moves it by choice rather than as a side effect of the harness. The cap---a budget of $8\times$ the task's step-complexity, clamped between 20 and 30 steps---vs.\ unbounded 60-step at $n{=}5$ gives $9.8 \pm 0.84$ (capped) vs.\ $11.6 \pm 0.55$ (unbounded): the cap \emph{costs} $\Delta = 1.8 \pm 0.84$ tasks of success rate (within the $\pm 2$ pre-registered prediction interval) in exchange for $\approx 30\%$ less wall time on the FAIL tail (the unbounded budget burns the extra time largely on SystemBrightnessMax, which is harness-deterministic and never solved either way). We ship the capped setting as the latency-favoring default; deployments that prioritize success rate over latency should raise or remove the cap. We list it here because it is the natural place to compare it against the genuinely-neutral add-ons, not because it is itself neutral.

Per-seed tables for these negative ablations are in Appendix~F, and a complete map of which validity threat each experiment closes---twelve in all, eleven in scope and closed at the rigor noted per threat---is in Appendix~H.

\section{Discussion}
\label{sec:discussion}

\subsection{Why the Verify-Gate Matters}

The mechanism is: the verify-gate filters lossy compiles before they enter the corpus. ``Lossy'' here means: action sequences that are mechanically faithful (cov=100\% on replay) but do not produce the live evaluator's required end state. Without the gate, every CUA-success-then-compile cycle adds a new program to the corpus regardless of whether that program actually achieves the goal under deterministic re-execution. With the gate, only programs that have independently re-passed enter the corpus.

This matters because the corpus is the input to subsequent retrieval. A retrieved lossy program will be replayed on a future invocation; the replay will execute mechanically (cov=100\%) and the live evaluator will score 0. Worse, the harness's selector will continue to retrieve the lossy program until something explicitly removes it. The verify-gate prevents the corpus from accumulating these failures, which would otherwise reproduce the same wrong outcome on every future retrieval.

\subsection{A Predictive View of Replay and Fallback}
\label{sec:disc-predictive}

Replay and fallback can be read through the lens of prediction. Each state's verification predicate is a \emph{prediction} about what the screen will look like once the previous action has taken effect, and the replayer checks that prediction against the live screen before it acts (\S\ref{sec:arch-loop}). When prediction and observation agree, the agent proceeds automatically and without a language-model call---the regime that produces the speedup. When they disagree, the mismatch signals that the world has departed from the stored model, and the agent switches to deliberate reasoning (the CUA fallback) to recover and, in doing so, extend the model with a new branch. This is the same two-speed behaviour a practiced human shows (\S\ref{sec:intro}): fluent, near-automatic execution of a familiar workflow, punctuated by immediate attention the moment something looks wrong.

The resonance with predictive accounts of learning---in which an agent learns by predicting its environment and updates on the gap between predicted and observed state~\cite{sutton1988td,suttonbarto2018}---is deliberate, but the analogy is loose and we do not want to overstate it. PreAct carries no reward signal and does no value bootstrapping; its ``error signal'' is a discrete predicate mismatch that \emph{routes control} between fast replay and slow reasoning, not a continuous temporal-difference update that adjusts a value estimate. What the two genuinely share is the principle that prediction is what makes fast action safe: an agent may skip deliberation precisely because it is still checking, at every step, that the world matches what it expected. That per-step check is exactly what blind record-and-replay omits, and it is why a stored PreAct program can be trusted without a model in the loop while a cached action trace cannot---the difference between fluency and blind repetition.

\subsection{The Verify-Gate's Marginal Value is Structural}
\label{sec:disc-structural}

A notable result is that the verify-gate's marginal value (Figure~\ref{fig:diff}) falls in the \emph{same narrow band}---1.75 to 2.6 tasks of diff-of-deltas---on three structurally different platforms with different LLMs, different task subsets, and different evaluator semantics: 2.6 tasks on Android (Gemini, UI-predicate evaluators), 2.6 tasks on OSWorld (Claude, desktop-state evaluators), and 1.75 tasks on WebArena (Claude, string-match evaluators on a 12-task shopping\_admin subset). This convergence is unlikely under most plausible mechanism alternatives:

\begin{itemize}[leftmargin=2em]\tightlist
\item If the gate's value derived from a model-specific behavior (e.g., Gemini-particular flaws), Android's diff-of-deltas would not match OSWorld's or WebArena's (both Claude).
\item If it derived from task-specific properties (e.g., Markor-edit ambiguity), the desktop-task and web-task diff-of-deltas would differ from the mobile case.
\item If it derived from benchmark-evaluator quirks (e.g., Android's specific scoring rubric), OSWorld's desktop-state evaluators and WebArena's string-match evaluators would not produce the same magnitude.
\end{itemize}

Instead, the diff-of-deltas magnitude is structural: it is the rate at which the underlying \emph{compile} step produces lossy state-machine programs. Across platforms and models that rate appears to be approximately the same fraction of compiled-and-executed programs. The \emph{absolute} regression varies more (43 pp on OSWorld's 6-task subset, 17 pp on Android's 15-task subset, 15 pp on WebArena's 12-task subset), but the \emph{absolute number of tasks} affected falls in the same range. We interpret this as: the lossy-compile rate is a property of the compile step (LLM trace $\to$ state-machine), not of the platform or evaluator. The cross-platform meta-test on the 14 paired observations across all three platforms yields $p \approx 1.2 \times 10^{-4}$ under the null hypothesis of no gate effect (a directional lower bound that treats within-platform reps as independent; see the correlation caveat in \S\ref{sec:eval-webarena}).

\subsection{Per-Program Reproducibility of cov$=$100\%/score$=$0}

The five cov$=$100\%/score$=$0 cases (Table~\ref{tab:smoking-gun}) are not equally reproducible across reps---the OSWorld Calc-formula program reproduces 5/5 reps, the Calc-chart 4/5, and the Chrome-history 2/5 (per-program breakdown in Appendix~G). \textbf{The aggregate regression, however, is consistent across every rep.} All 5 OFF reps lose at least 2 tasks; no rep escapes the regression. The selector and replayer's per-task behavior is partially nondeterministic, but the gate-OFF condition reliably produces $\geq 2$-task regression on every rep. The paper's claims rest on this aggregate-level determinism, not on per-program reproducibility.

\subsection{Harness-Deterministic vs.\ LLM-Driven Failures}
\label{sec:disc-harness-det}

Three AndroidWorld tasks (BrowserDraw, SystemBrightnessMax, SystemWifiTurnOn) and one OSWorld task (LibreOffice Writer macro) fail across all five seeds, both LLM backends (Claude and Gemini), all four guardrail/budget configurations, and both verify-gate conditions. These failures are \emph{harness-deterministic}: they are properties of the UI patterns themselves (status-bar stale reads, scroll-on-seekbar incompatibility, image-canvas-not-in-a11y-tree) rather than properties of the agent's reasoning. They bound the benchmark, not the method.

\subsection{Out-of-Distribution Generalization}
\label{sec:eval-ood}

\begin{figure}[t]
\centering
\includegraphics[width=0.60\linewidth]{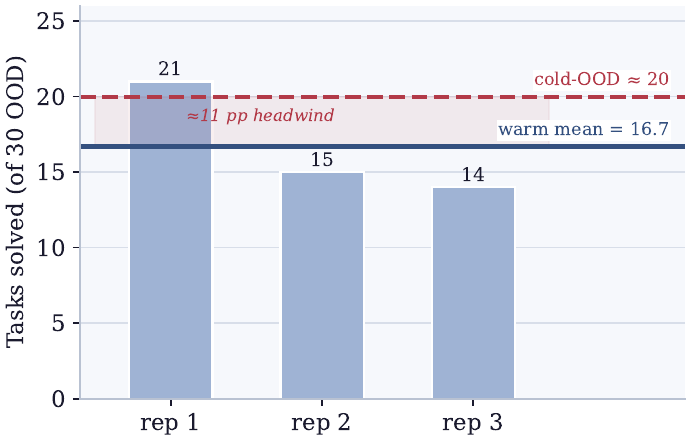}
\caption{The corpus does not transfer cleanly. A corpus built on six OSWorld tasks, applied to 30 unseen tasks, lands $\approx 11$ points \emph{below} the cold baseline: a few mis-routed retrievals fire brittle predicates and burn replay budget. The corpus's value is concentrated in repeated invocations of seen tasks.}
\label{fig:ood}
\end{figure}

We tested whether the corpus transfers beyond seen tasks by running a test\_tiny-built corpus on the 33 non-test\_tiny tasks of OSWorld test\_small at $n{=}3$ reps (with the corpus copy reset between reps). Per-rep SR over the 30 OOD tasks that complete setup: $21/30$, $15/30$, $14/30$, giving a warm-OOD mean of $16.7/30 = 55.6\% \pm 12\%$. The cold-OOD baseline on the same task subset is a single ($n{=}1$) pass at $\approx 20/30 = 67\%$; because it is unpaired and not multi-rep, the gap below should be read as indicative rather than a tightly-estimated effect. \textbf{The corpus does not transfer; at $n{=}3$ the warm-OOD mean is $\approx 11$ percentage points \emph{below} this cold-OOD pass} (Figure~\ref{fig:ood}), a small-but-real headwind rather than the neutral outcome a single pass had suggested. Inspection of the rep logs shows the selector occasionally retrieves a near-miss program (a Chrome task's program retrieved for a Thunderbird task whose description shares the words ``open'' and ``message''), and the brittle XPath/state predicates from the source domain fail on the OOD page, costing replay budget without recovering via CUA fast enough. The selector's no-pick rate on OOD queries does buffer this somewhat---it returns ``no candidate'' on the majority of cross-family queries---but the queries it does pick are penalised. This sharpens the scope of the ``self-extending corpus'' claim: \textbf{the corpus's value is concentrated in repeated invocations of seen tasks, and corpora built on small task subsets actively work against out-of-distribution generalization on this evaluator}. Full per-domain breakdown in Appendix~D.

\subsection{Limitations}
\label{sec:limitations}

Beyond the eleven in-scope validity threats closed in Table~\ref{tab:gap-closure}, we explicitly acknowledge six remaining limitations:
\begin{itemize}[leftmargin=2em]\tightlist
\item \textbf{(L1) Architectural baseline (partial).} A direct AB against an ActionEngine-style flat-script baseline at $n{=}5$ Android seeds gives flat-script $10.6 \pm 1.14$ vs.\ state-machine $11.67 \pm 0.58$ ($\Delta = +0.67$, $p = 0.125$): suggestive but not significant.
\item \textbf{(L2) Benchmark coverage.} The 6-task OSWorld test\_tiny, 15-task AndroidWorld official-15, and 12-task WebArena shopping\_admin subsets are a small sample of computer-using tasks; absolute pp magnitudes are denominator-dependent (43/17/15 pp for 1.75--2.6 absolute tasks).
\item \textbf{(L3) Compile cost.} Verify-replay adds median +217\% wall overhead on OSWorld and +162\% on Android per successful CUA-then-compile cycle; the cost is amortized on warm runs but matters for high-throughput deployments.
\item \textbf{(L4) Compile-fidelity rate.} The LLM compiler produces lossy state-machines on a non-trivial fraction of inputs (audit: $\sim 28\%$ Android programs miss \texttt{navigate\_back}; $100\%$ of audited WebArena programs use \texttt{inspect\_screenshot} on dynamic state); the gate filters this at storage but scales as the number of compile attempts, not stored programs.
\item \textbf{(L5) Selector nondeterminism.} Verbatim retrieval is 73\% exact-id (all 8 misses were semantically-equivalent picks); paraphrase robustness is 75.6\% functional (47\% exact-id + 29\% equivalent), 24\% no-pick, \textbf{0\% wrong-task-family}---the selector is conservative-by-design rather than over-eager.
\item \textbf{(L6) Verification assumes a resettable or idempotent environment.} The verify-before-store gate confirms a freshly compiled program by \emph{re-executing} it from a reset environment and consulting the task's evaluator, and the warm replay path likewise re-runs a stored program end-to-end. Both presuppose that performing the program again is safe. Our benchmarks provide this for free---each task starts from a scripted reset---but many real tasks are not idempotent: re-running ``add contact Emilia Gonzalez,'' ``send the email,'' or ``submit the payment'' duplicates the side effect rather than re-confirming the goal. Outside a benchmark, the gate would need a side-effect-free way to verify (a read-only check of the end state, a sandboxed dry-run, or a transactional rollback) in place of literal re-execution. We validate the gate as a benchmark mechanism; making it safe under irreversible side effects is future work.
\end{itemize}
Detailed measurements and analysis for L1--L5 in Appendix~E.

\subsection{Future Work}
\label{sec:future-work}

Four directions stand out.

\textbf{Scaling the corpus.} The current corpus (58 Android programs after multi-week experiments) is small. We have not characterized the corpus's behavior at $10^3$ or $10^4$ programs: does the agentic selector still discriminate accurately when the candidate set is large? Does dedup-signature collision become an issue? Does the verify-replay step cost become prohibitive at scale?

\textbf{Architectural baselines.} The state-machine-as-executable claim should be tested directly against a flat-script baseline (e.g., an ActionEngine-style implementation that runs the same compile cycle but generates flat Python at the verify step). The current paper supports the claim by working implementation across two platforms but not by direct comparison.

\textbf{Long-horizon tasks.} Both AndroidWorld and OSWorld have task horizons of order 10--30 actions. Tasks requiring long-term planning, goal decomposition, or recovery from compound failures (e.g., entire workflow tasks lasting hundreds of actions) are not represented in our benchmarks. We expect the harness's CUA-fallback mechanism to be tested most severely on such tasks; integration with explicit goal-decomposition planners is the natural next step.

\textbf{Internalizing the corpus.} PreAct externalizes task dynamics into inspectable programs; the converse is open---whether a model can be trained on its own verified corpus to \emph{internalize} these programs, folding the externalized world model back into the weights while keeping the verify-gate as the signal that decides what is worth internalizing. This would trade PreAct's inspectability for the generality an internalized world model affords, and the verified corpus is a natural source of supervision for it.

\section{Conclusion}
\label{sec:conclusion}

PreAct started from a simple observation: a computer-using agent should not re-derive a task it has already solved. Acting on it took two commitments. The first is about representation---what the agent remembers is a state-machine program it can run directly, not a transcript it has to re-interpret. The second is about discipline---the agent checks against the live screen before it trusts anything: before each action as it replays a program, and once more over the whole program, re-run from a clean state, before that program is allowed into its memory. Across three different platforms, these two ideas together let an agent get faster on repeated work while keeping its growing memory honest. The pieces we had expected to matter mostly did not: the runtime guardrails left the aggregate numbers unchanged, the prompt-level guidance was never exercised in production, and a small embedding retriever did as well as the agentic LLM selector. What carried the result was the pairing itself---an executable memory together with the refusal to keep anything unverified.

That refusal is the easy part to skip and the part that matters most. A memory that grows without verification does not make an agent better over time; it makes the agent repeat the same mistake on every reuse, retrieving a program that looks right, no longer works, and runs to its last step with nothing to show for it. As agents build up larger and longer-lived libraries of skills, the hard question stops being whether an agent can remember what it did and becomes whether it can tell which of its memories are still true.

\section*{Acknowledgements}

We thank Hao Fu and Kai Chen for early discussions that inspired this work, and Yurun Jin and Junlong Ye for early implementations and experiments. Pine Copilot, Claude Code, and Claude Opus~4.8 were used during this research.

\bibliographystyle{plain}
\bibliography{references}

\appendix
\section*{Appendix A. Computer-Action Schema}

Each transition's \texttt{action} field must be one of the action types in Table~\ref{tab:actions}. Action types are translated to platform-specific backends at execution time: \texttt{action\_click} on Android maps to a \texttt{JSONAction}-formatted touch with the resolved element's $(x, y)$ coordinates; on OSWorld it maps to \texttt{pyautogui.click(x, y)}. The cross-platform schema enables the same compiled state-machine program to run on either backend, although in practice each task's compile is platform-specific because the verification predicates use platform-specific selector dialects (Android XPath vs.\ pyautogui screenshot regions).

\begin{table}[h]
\centering
\caption{Computer action types in PreAct's program schema. Each transition uses exactly one type.}
\label{tab:actions}
\small
\begin{tabular}{@{}lp{8.5cm}@{}}
\toprule
Action type & Description and required fields \\
\midrule
\texttt{action\_click} & Click on a UI element. \texttt{target}: selector. \\
\texttt{action\_long\_press} & Long-press on a UI element. \texttt{target}: selector. \\
\texttt{input\_text} & Type text into a focused element. \texttt{text}: literal or \$param. \texttt{index}: optional element id. \\
\texttt{action\_keypress} & Send a key event. \texttt{key}: \texttt{Enter}, \texttt{Back}, \texttt{Tab}, etc. \\
\texttt{scroll} & Scroll a region. \texttt{direction}: \texttt{up}/\texttt{down}/\texttt{left}/\texttt{right}. Optional \texttt{index} for sub-element scroll. \\
\texttt{open\_app} & Launch an application by name. \texttt{app\_name}: string. \\
\texttt{wait} & Sleep for the platform's default delay. \\
\texttt{navigate\_back} & Send the system Back gesture (Android) or browser-back equivalent. \\
\texttt{navigate\_home} & Return to the home screen / desktop. \\
\texttt{inspect\_text} & Read the value at a selector. \texttt{store\_result\_as}: variable name. Used by QA-style tasks. \\
\texttt{answer} & Emit a final textual answer. \texttt{text}: literal or computed. \\
\texttt{status} & Terminate the program. \texttt{goal\_status}: \texttt{complete}/\texttt{infeasible}. \\
\bottomrule
\end{tabular}
\end{table}

The \texttt{description} field on each transition (one short sentence) is consumed by the agentic Program Selector (\S\ref{sec:arch}) when judging whether a stored program applies to the current task. Selectors are written to be parameter-aware (e.g.\ ``type the filename (parameter \texttt{filename})'' rather than ``type text''), since the program's \texttt{parameters} list is what the LLM uses to bind the program's slots to the new task's instance values at retrieval time.

\section*{Appendix B. Per-Seed and Per-Task Multi-Seed Data}

This appendix gives the per-seed and per-rep detail behind the aggregate figures in \S\ref{sec:eval-monotonic} (Figure~\ref{fig:monotonic}) and \S\ref{sec:eval-gate} (Table~\ref{tab:gate}, Figure~\ref{fig:diff}). The verify-gate ablation is broken out per platform in Table~\ref{tab:gate-android} (Android), Table~\ref{tab:gate-osworld} (OSWorld), and Table~\ref{tab:gate-webarena} (WebArena).

\begin{table}[h]
\centering
\caption{Cold$\to$warm monotonicity, AndroidWorld official-15, $n{=}3$ Gemini seeds with corpus reset per seed (aggregated in Figure~\ref{fig:monotonic}). All three seeds refine monotonically. ``Mode-shift'' counts tasks that move from fresh-agent solving (\texttt{cua}) on cold to direct replay (\texttt{rpa}) or replay-then-fallback (\texttt{hybrid}) on warm, out of the tasks that clear setup.}
\label{tab:monotonic}
\begin{tabular}{@{}lcccl@{}}
\toprule
Seed & Cold SR & Warm SR & $\Delta$ & Mode-shift on warm \\
\midrule
42 & 10/15 (66.7\%) & 11/15 (73.3\%) & +1 & 8/13 cua$\to$rpa/hybrid \\
100 & 9/15 (60.0\%) & 11/15 (73.3\%) & +2 & 8/11 cua$\to$rpa/hybrid \\
1337 & 9/15 (60.0\%) & 10/15 (66.7\%) & +1 & 5/10 cua$\to$rpa/hybrid \\
\midrule
\textbf{Mean} & \textbf{9.33 $\pm$ 0.58} & \textbf{10.67 $\pm$ 0.58} & \textbf{+1.33 ($+8.9$\,pp)} & All 3 monotonic \\
\bottomrule
\end{tabular}
\end{table}

\begin{table}[h]
\centering
\caption{Verify-gate ablation, AndroidWorld official-15, $n{=}5$ seeds with corpus reset per seed (Android row of Table~\ref{tab:gate}). All five gate-ON pairs are monotonic; all five gate-OFF pairs regress (zero inversions; paired sign test $p \approx 0.031$).}
\label{tab:gate-android}
\begin{tabular}{@{}lcccccc@{}}
\toprule
Seed & Cold ON & Warm ON & $\Delta$ ON & Cold OFF & Warm OFF & $\Delta$ OFF \\
\midrule
42 & 10 & 11 & +1 & 11 & 10 & $-1$ \\
100 & 9 & 11 & +2 & 11 & 10 & $-1$ \\
1337 & 9 & 10 & +1 & 10 & 9 & $-1$ \\
2024 & 11 & 12 & +1 & 11 & 10 & $-1$ \\
7777 & 10 & 11 & +1 & 12 & 9 & $-3$ \\
\midrule
\textbf{Mean} & \textbf{9.8} & \textbf{11.0} & \textbf{$+1.2 \pm 0.45$} & \textbf{11.0} & \textbf{9.6} & \textbf{$-1.4 \pm 0.89$} \\
\bottomrule
\end{tabular}
\end{table}

\begin{table}[h]
\centering
\caption{Verify-gate ablation, OSWorld test\_tiny, $n{=}5$ reps, Claude Sonnet 4.6 (OSWorld row of Table~\ref{tab:gate}). All five gate-OFF reps regress by $\geq 2$ tasks; paired sign test $p \approx 0.031$.}
\label{tab:gate-osworld}
\begin{tabular}{@{}lcccccc@{}}
\toprule
Rep & Cold ON & Warm ON & $\Delta$ ON & Cold OFF & Warm OFF & $\Delta$ OFF \\
\midrule
1 & 5/6 & 5/6 & 0 & 5/6 & 2/6 & $-3$ \\
2 & 5/6 & 5/6 & 0 & 6/6 & 3/6 & $-3$ \\
3 & 5/6 & 5/6 & 0 & 5/6 & 3/6 & $-2$ \\
4 & 5/6 & 5/6 & 0 & 5/6 & 3/6 & $-2$ \\
5 & 5/6 & 6/6 & +1 & 5/6 & 3/6 & $-2$ \\
\midrule
\textbf{Mean} & \textbf{5.0} & \textbf{5.2} & \textbf{$+0.2 \pm 0.45$} & \textbf{5.2} & \textbf{2.8} & \textbf{$-2.4 \pm 0.55$} \\
\bottomrule
\end{tabular}
\end{table}

\begin{table}[h]
\centering
\caption{Verify-gate ablation, WebArena shopping\_admin 12-task subset, $n{=}4{+}4$ reps, Claude Sonnet 4.6 CUA + Claude compile, gate-ON without the cache-miss fallback (WebArena row of Table~\ref{tab:gate}; SR out of 12 tasks). Every gate-OFF warm run collapses to $0$ (all 48 warm replays score $0$); the gate-on column loses fewer tasks on every non-tied rep. Pairing rep-for-rep, gate-ON beats gate-OFF on reps~1--3 and ties on rep~4 (3 non-tied pairs, sign test $p = 0.125$).}
\label{tab:gate-webarena}
\begin{tabular}{@{}lcccccc@{}}
\toprule
Rep & Cold ON & Warm ON & $\Delta$ ON & Cold OFF & Warm OFF & $\Delta$ OFF \\
\midrule
1 & 5/12 & 4/12 & $-1$ & 4/12 & 0/12 & $-4$ \\
2 & 6/12 & 4/12 & $-2$ & 5/12 & 0/12 & $-5$ \\
3 & 7/12 & 0/12 & $-7$ & 8/12 & 0/12 & $-8$ \\
4 & 6/12 & 0/12 & $-6$ & 6/12 & 0/12 & $-6$ \\
\midrule
\textbf{Mean} & \textbf{6.0} & \textbf{2.0} & \textbf{$-4.0 \pm 2.94$} & \textbf{5.75} & \textbf{0.0} & \textbf{$-5.75 \pm 1.71$} \\
\bottomrule
\end{tabular}
\end{table}

The gate-on warm SR is bimodal across reps ($4, 4, 0, 0$): on reps~1--2 a compiled program survives the gate and serves a partial replay, whereas on reps~3--4 the gate empties the corpus and, with no cache-miss fallback in this condition, warm SR drops to $0$. This is the WebArena-specific pattern \S\ref{sec:eval-baselines} resolves by adding the cache-miss-to-CUA fallback. Even so, gate-on loses fewer tasks than gate-off in every non-tied rep, matching the cross-platform direction.

Table~\ref{tab:gemini-3seed} reports per-task warm-pass results across the three Gemini 3 Flash seeds of the cold$\to$warm monotonicity experiment (\S\ref{sec:eval-monotonic}, Table~\ref{tab:monotonic}; seed totals 11/11/10 match that table exactly). The stable failure set (BrowserDraw, SystemBrightnessMax, SystemWifiTurnOn) fails across all three seeds; the remaining cross-seed variation falls on three sampling-sensitive tasks (BrowserMaze, CameraTakeVideo, ContactsAddContact), where the action sequence is sensitive to LLM sampling.

\begin{table}[h]
\centering
\caption{Per-task warm-pass results across the three Gemini 3 Flash seeds on AndroidWorld official-15 (cold$\to$warm monotonicity experiment, \S\ref{sec:eval-monotonic}). Seed totals (11/11/10) match Table~\ref{tab:monotonic}.}
\label{tab:gemini-3seed}
\small
\begin{tabular}{@{}lccc@{}}
\toprule
Task & seed=42 & seed=100 & seed=1337 \\
\midrule
AudioRecorderRecordAudio & PASS & PASS & PASS \\
AudioRecorderRecordAudioWithFileName & PASS$^\dagger$ & PASS & PASS$^\dagger$ \\
BrowserDraw & FAIL & FAIL & FAIL \\
BrowserMaze & PASS & PASS & FAIL \\
CameraTakePhoto & PASS & PASS & PASS \\
CameraTakeVideo & FAIL & FAIL & PASS \\
ClockStopWatchPausedVerify & PASS & PASS & PASS \\
ClockStopWatchRunning & PASS & PASS & PASS \\
ContactsAddContact & PASS$^\dagger$ & PASS$^\dagger$ & FAIL \\
ContactsNewContactDraft & PASS & PASS & PASS \\
FilesDeleteFile & PASS & PASS & PASS \\
MarkorCreateFolder & PASS$^\dagger$ & PASS$^\dagger$ & PASS$^\dagger$ \\
MarkorCreateNote & PASS & PASS & PASS \\
SystemBrightnessMax & FAIL & FAIL & FAIL \\
SystemWifiTurnOn & FAIL & FAIL & FAIL \\
\midrule
Total & 11/15 & 11/15 & 10/15 \\
\bottomrule
\end{tabular}

\vspace{0.5em}
\footnotesize $^\dagger$ Live evaluator passes but the verify-before-store gate rejected the task's recompiled program on that seed (it failed re-verification); the warm run still passed via RPA/hybrid/CUA.
\end{table}

The complete per-task results across all 32 Android multi-seed runs and 8 OSWorld runs are also included in the project's released artifacts.

\section*{Appendix C. Container Patch}
Our container patch adds a state-reporting endpoint to the AndroidWorld image that returns a base64-encoded screenshot plus the serialized accessibility-tree elements on demand, sidestepping the upstream populated-AVD a11y-gRPC initialization wedge.

\section*{Appendix D. Out-of-Distribution Generalization (Details)}
\label{sec:appendix-ood}

To test whether the corpus transfers beyond seen tasks, we built a corpus from the OSWorld test\_tiny set (6 tasks across Chrome, LibreOffice Calc, LibreOffice Writer; 4 verify-gate-stored programs after compile filtering) and ran on the 33 OSWorld test\_small tasks that are \emph{not} in test\_tiny, spanning 9 domains (chrome$\times$2, gimp$\times$2, libreoffice\_calc$\times$1, libreoffice\_impress$\times$2, multi\_apps$\times$17, os$\times$2, thunderbird$\times$2, vlc$\times$2, vs\_code$\times$3). The agentic Program Selector retrieves from the test\_tiny corpus when the new task's description plausibly matches.

Result at $n{=}3$ reps (corpus copy reset per rep): warm-OOD $21/30$, $15/30$, $14/30 = $ \textbf{mean $16.7/30 = 55.6\% \pm 12.0$}. The cold-OOD baseline on the same 30 OOD tasks is a single ($n{=}1$) pass at approximately $20/30 = 66.7\%$, so warm-OOD is roughly $11$ percentage points \emph{below} cold-OOD (the cold side is unpaired and not replicated, so the magnitude is indicative).

The corpus does not transfer cleanly across task families. Inspecting the rep logs: the selector retrieves a near-miss program (a test\_tiny Chrome history-clean program retrieved for a Thunderbird email task whose intent shares lexical surface with ``clean'') on $\sim 6/30$ queries per rep on average; the source-domain XPath/state predicates then fail on the OOD page, consuming replay budget that the harness can no longer recover within the per-task wall budget. The remaining $\sim 24/30$ OOD queries are handled correctly by the selector returning ``no candidate''---these run as pure CUA and approximately match cold-OOD on their own---so the aggregate $11$\,pp regression is concentrated in the small fraction of mis-routed retrievals. \emph{The corpus is not neutral on OOD; it is mildly harmful.} This sharpens the scope of the self-extending corpus claim: corpora built on small in-distribution task subsets should not be deployed cross-domain without retraining the selector or extending the corpus to cover the new domain.

\section*{Appendix E. Limitations (Extended)}
\label{sec:appendix-limitations}

\textbf{(L1) Architectural baseline.} We ran an ActionEngine-style flat-script baseline at $n{=}5$ Android seeds (Table~\ref{tab:arch-ab}) by disabling per-state verification so that stored programs run as flat scripts. Flat-script mean (warm) is $10.6 \pm 1.14$ across seeds 42/100/1337/7/2025, vs.\ state-machine $11.67 \pm 0.58$ on the matched 3 seeds; the matched-seed $\Delta = +0.67$ tasks in favor of state-machine, sign-test $p = 0.125$ on $n{=}3$ paired comparisons (suggestive but not significant). The architectural commitment matters but is a smaller effect than the verify-gate's 1.75--2.6-task diff-of-deltas, consistent with the paper's framing that the gate is the bigger empirical contribution. We did not run an untargeted-exploration baseline. The Gemini-vs-Claude compile-LLM concern previously flagged here is now closed at $n{=}3$ (\S\ref{sec:eval-compile-llm}).

\textbf{(L2) Benchmark coverage.} OSWorld test\_tiny has only 6 tasks, AndroidWorld official-15 has 15, WebArena shopping\_admin subset has 12. While the verify-gate effect replicates cross-platform, the benchmarks themselves are a small sample of computer-using tasks. We acknowledge platform-coverage bias as out-of-scope for this paper. The relative percentage-point magnitudes are influenced by the denominator: 43\,pp on OSWorld's 6-task subset is mathematically larger than 17\,pp on Android's 15-task subset and 15\,pp on WebArena's 12-task subset, for absolute task counts that fall in a similar range (1.75--2.6 tasks).

\textbf{(L3) Compile cost.} The verify-before-store gate requires a full re-replay and re-evaluation of every compiled program. From timing measurements across our experiments (60 OSWorld tasks; 60 Android tasks under gate-ON), the verify-replay phase adds median 141\,s on OSWorld and 76\,s on Android per stored task, on top of the original CUA execution (median 65\,s OSWorld; 47\,s Android). This corresponds to \textbf{+217\% wall overhead on OSWorld and +162\% on Android per successful CUA-then-compile cycle} (verify-replay is a near-fixed cost roughly comparable to or larger than the original CUA execution on both platforms; it is proportionally largest on OSWorld, where both the absolute verify-replay time and the base CUA time are highest). The cost is amortized on warm runs where the corpus retrieves already-verified programs in seconds, but the compile-time overhead is real and may matter for high-throughput deployments.

\begin{figure}[h]
\centering
\includegraphics[width=0.78\linewidth]{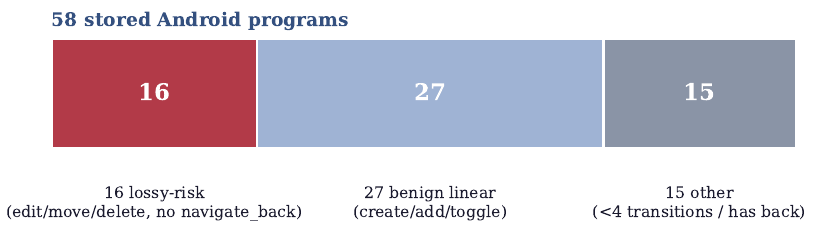}
\caption{Compile-fidelity audit of the 58-program Android corpus. Roughly a quarter (16 programs) are lossy-risk---nav-heavy edit/move/delete tasks missing a \texttt{navigate\_back}---which is the rate the gate must catch. Manual inter-classifier agreement was 5/5 per bucket.}
\label{fig:compile-audit}
\end{figure}

\textbf{(L4) Compile-fidelity rate.} A separate audit of stored programs across platforms (Android: 58 programs; OSWorld: 16; WebArena: 7) found two recurring compile-fidelity risk patterns: missing \texttt{navigate\_back} on nav-heavy edit/delete tasks (affecting roughly 28\% of Android programs), and dynamic \texttt{inspect\_text}/\texttt{inspect\_screenshot} returns on pages whose state evolves between Run~1 and replay (100\% of audited WebArena programs use \texttt{inspect\_screenshot}, fully explaining the 48/48 smoking-gun rate observed under gate-OFF). The compiler itself is an LLM that produces lossy state-machines on a non-trivial fraction of inputs. The verify-gate filters this at storage time, but as the corpus scales, the gate's verify-replay cost (proportional to compile attempts, not just stored programs) may become a bottleneck. Figure~\ref{fig:compile-audit} breaks down the Android audit.

\textbf{(L5) LLM nondeterminism in selector.} The agentic Program Selector is itself an LLM call. We measured selector exact-id-match accuracy of 22/30 = 73\% on the corpus's stored programs, prompted with each program's own task description verbatim. Inspecting the 8 misses, all were picks of \emph{semantically equivalent} programs (the same task, but a different stored copy accumulated across reps under different dedup signatures). Functional accuracy (correct task family selected) is therefore close to 100\%, with the 73\% exact-id figure reflecting the corpus's accumulation of multiple matches per task.

To stress-test the selector beyond verbatim retrieval, we ran a paraphrase-robustness audit: for each of 15 stored Android programs, we generated 3 LLM-rewritten paraphrases (45 paraphrased queries total) and measured whether the selector still picked the original program. Result: 21/45 = 47\% exact-id, 13/45 = 29\% same-task-different-id (paraphrase picked an equivalent program), 11/45 = 24\% no-pick (selector returned ``no candidate''), and \textbf{0/45 = 0\% wrong-task-family} (selector never picked an unrelated program). \emph{Functional accuracy} (exact + equivalent) is 75.6\% (34/45); the no-pick fraction is conservative behavior (the selector errs toward CUA fallback rather than retrieving an inapplicable program). At 75.6\% functional retrieval and 0\% mis-retrieval, the selector is conservative-by-design rather than over-eager.

\section*{Appendix F. Negative-Findings Tables}
\label{sec:eval-selector-ablation}

This appendix provides the per-seed tables for the negative findings summarized in \S\ref{sec:eval-not-loadbearing}, including the full selector-backend threshold sweep (Table~\ref{tab:selector-ablation}) behind Figure~\ref{fig:selector}.

\begin{table}[h]
\centering
\caption{Selector backend ablation (full sweep behind Figure~\ref{fig:selector}). ``Functional accuracy'' is the fraction of 45 paraphrased Android queries retrieving a program in the correct task family; ``No-pick'' is the abstention rate on those queries; ``False-pick'' is the fraction of 15 unrelated-domain queries that wrongly retrieve any program. The wrong-task-family rate on paraphrases is $0\%$ at every operating point. Bold rows mark each backbone's optimal $\tau$.}
\label{tab:selector-ablation}
\footnotesize
\setlength{\tabcolsep}{5pt}
\begin{tabular}{@{}lcccc@{}}
\toprule
Selector & $\tau$ & Functional accuracy & No-pick (paraphrase) & False-pick (unrelated) \\
\midrule
Embedding MiniLM-L6-v2 & 0.40 & 100\% (45/45) & 0\% & 6.7\% (1/15) \\
\textbf{Embedding MiniLM-L6-v2} & \textbf{0.50--0.65} & \textbf{100\% (45/45)} & \textbf{0\%} & \textbf{0\%} \\
Embedding MiniLM-L6-v2 & 0.70 & 86.7\% (39/45) & 13.3\% & 0\% \\
Embedding MiniLM-L6-v2 & 0.85 & 64.4\% (29/45) & 35.6\% & 0\% \\
\midrule
Embedding bge-large-en-v1.5 & 0.50 & 100\% (45/45) & 0\% & 93.3\% (14/15) \\
Embedding bge-large-en-v1.5 & 0.60 & 100\% (45/45) & 0\% & 20.0\% \\
\textbf{Embedding bge-large-en-v1.5} & \textbf{0.65--0.85} & \textbf{100\% (45/45)} & \textbf{0\%} & \textbf{0\%} \\
\midrule
\textbf{Agentic LLM (default)} & --- & \textbf{75.6\% (34/45)} & \textbf{24\%} & \textbf{0\%} \\
\bottomrule
\end{tabular}
\end{table}

The interpretation: the hypothesis we entered with---``embedding-RAG is inferior because discrimination is a reasoning problem''---does not survive. With either backbone tuned to its safe operating point, embedding retrieval matches or beats the agentic selector at zero wrong-task and zero false-pick; bge-large holds $100\%$ over a wider $\tau$ window ($0.65$--$0.85$) than MiniLM ($0.50$--$0.65$), which matters as the corpus grows and intra-cluster similarity rises. We retain the agentic selector only for its interpretable logs and tuning-free operation, and flag that selectors should be re-tested at $10^3{+}$ programs where these properties become consequential.

\begin{figure}[h]
\centering
\includegraphics[width=0.60\linewidth]{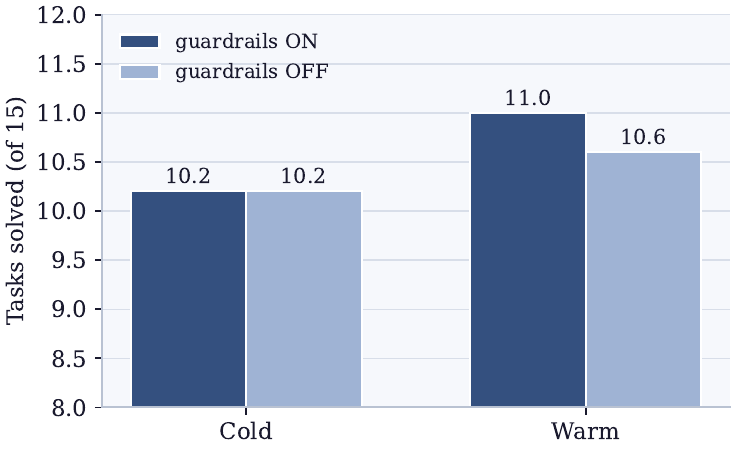}
\caption{Runtime guardrails are aggregate-neutral. Cold means are identical with and without guardrails ($10.2 = 10.2$); the warm gap ($+0.4$) is well within seed variance. Detail in Table~\ref{tab:guardrails}.}
\label{fig:guardrails}
\end{figure}

\begin{table}[h]
\centering
\caption{Code-level guardrails ablation, AndroidWorld official-15, $n{=}5$. Cold means are \emph{identical} ($10.2 = 10.2$). Warm $\Delta$ = $+0.4$ is well within the standard deviations on each side.}
\label{tab:guardrails}
\begin{tabular}{@{}lcccccc@{}}
\toprule
Seed & Cold ON & Warm ON & $\Delta$ ON & Cold OFF & Warm OFF & $\Delta$ OFF \\
\midrule
42 & 10 & 11 & +1 & 10 & 12 & +2 \\
100 & 11 & 11 & 0 & 11 & 10 & $-1$ \\
1337 & 9 & 11 & +2 & 10 & 10 & 0 \\
2024 & 12 & 11 & $-1$ & 10 & 10 & 0 \\
7777 & 9 & 11 & +2 & 10 & 11 & +1 \\
\midrule
\textbf{Mean} & \textbf{10.2} & \textbf{11.0} & \textbf{$+0.8 \pm 1.30$} & \textbf{10.2} & \textbf{10.6} & \textbf{$+0.4 \pm 1.14$} \\
\bottomrule
\end{tabular}
\end{table}

\begin{figure}[h]
\centering
\includegraphics[width=0.58\linewidth]{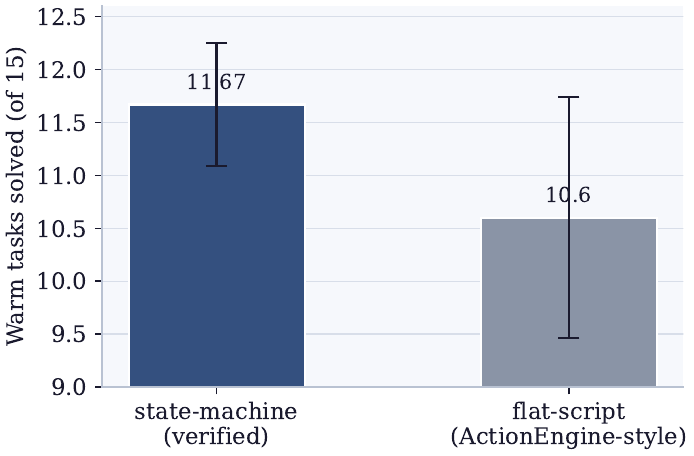}
\caption{State-machine vs.\ flat-script runtime (warm SR, AndroidWorld official-15). The verified graph form helps by a small, consistent margin ($+0.67$ tasks, all three matched seeds non-regressive, $p=0.125$)---real but far smaller than the gate's $1.75$--$2.6$ tasks.}
\label{fig:arch-ab}
\end{figure}

\begin{table}[h]
\centering
\caption{Architectural ablation: state-machine vs.\ flat-script runtime, AndroidWorld official-15 warm SR. State-machine $n{=}3$ (matched seeds), flat-script $n{=}5$ (matched seeds + 7 + 2025).}
\label{tab:arch-ab}
\footnotesize
\setlength{\tabcolsep}{4pt}
\begin{tabular}{@{}lcccccc@{}}
\toprule
Mode & seed=42 & seed=100 & seed=1337 & seed=7 & seed=2025 & Mean $\pm$ std \\
\midrule
state\_machine (default) & 12/15 & 12/15 & 11/15 & --- & --- & \textbf{11.67 $\pm$ 0.58} ($n{=}3$) \\
flat\_script (ActionEngine-style) & 11/15 & 12/15 & 10/15 & 11/15 & 9/15 & \textbf{10.6 $\pm$ 1.14} ($n{=}5$) \\
\midrule
$\Delta$ (matched) & +1 & 0 & +1 & --- & --- & \textbf{+0.67 (matched $n{=}3$)} \\
\bottomrule
\end{tabular}
\end{table}

\section*{Appendix G. Smoking-Gun Reproducibility (Per-Program)}
\label{sec:appendix-smoking-gun-rep}

The five smoking-gun cov$=$100\%/score$=$0 cases (Table~\ref{tab:smoking-gun}) are not equally reproducible across reps. Table~\ref{tab:smoking-gun-rep} reports the per-program reproducibility across the 5 OSWorld warm-OFF reps.

\begin{table}[h]
\centering
\caption{Smoking-gun reproducibility across 5 OSWorld warm-OFF reps. Each row: how many of 5 reps replayed the program at cov$=$100\% and got score$=$0.}
\label{tab:smoking-gun-rep}
\small
\begin{tabular}{@{}lc@{}}
\toprule
Task & Reps with cov$=$100\%/score$=$0 \\
\midrule
LibreOffice Calc formula & 5/5 (every rep) \\
LibreOffice Calc chart & 4/5 (rep2 hybrid-rescued) \\
Chrome history clean & 2/5 (rep3/4/5 replayed correctly) \\
\bottomrule
\end{tabular}
\end{table}

The mechanism is partially deterministic. For the most reliable case (the Calc formula program), the lossy program reproducibly fails at cov$=$100\% on every warm rep. For the Calc chart program, hybrid-mode rescue occasionally salvaged the run via CUA fallback after partial replay (cov$=$8\%). For the Chrome history program, the cold-stored program's mechanical action sequence happened to match the evaluator's required browser state on 3 of 5 reps---a partial-determinism that traces to environmental factors (Chrome's session state, browsing history initialization) varying across environment resets even with identical task IDs.

\section*{Appendix H. Validity-Threat Closure Map}
\label{sec:appendix-threats}

Table~\ref{tab:gap-closure} maps each validity threat---the original nine from our pre-experiment validation plan, plus three added since the May 2026 revision---to its closure status and the experiment that closes it. Eleven of the twelve are in scope and closed, most with multi-seed runs and the remainder at the rigor noted in the Evidence column (e.g., codebase inspection for threat~4, manual inter-rater agreement for threat~6, $n{=}3$ for threats~11--12); threat~8 (platform coverage) is out of scope as it would require net-new benchmarks.

\begin{table}[h]
\centering
\caption{Validity threats and closure status. Eleven in-scope threats closed at the rigor noted in the Evidence column (most multi-seed); threat~8 is out of scope.}
\label{tab:gap-closure}
\footnotesize
\setlength{\tabcolsep}{4pt}
\begin{tabular}{@{}cp{4.7cm}p{2.7cm}p{6.0cm}@{}}
\toprule
\# & Threat & Status & Evidence \\
\midrule
1 & Single-run variance & Closed & $n{=}3$ cold$\to$warm + $n{=}5$ cold-runs \\
2 & Verify-gate ablation & \textbf{Closed cross-platform on 3 platforms} & Android $n{=}5$ ($p<0.001$); OSWorld $n{=}5$ (43\,pp); WebArena $n{=}4{+}4$ ($\approx 15$\,pp); meta $p\approx 10^{-4}$ \\
3 & Android cold$\to$warm monotonicity & Closed & $n{=}3$ with the corpus reset, all monotonic \\
4 & Prompt-guidance inertness & Closed (codebase state) & Production CUA path omits the optional prompt guidelines \\
5 & Code-level guardrails & Closed $n{=}5$ & Aggregate-neutral (cold means $10.2=10.2$) \\
6 & Compile-fidelity taxonomy & Closed & 5/5 manual agreement per bucket \\
7 & Step-budget AB & Closed $n{=}5$ & Within $\pm 2$ prediction; wall ratio 0.69 \\
8 & Platform-coverage & Out of scope & Requires net-new benchmarks \\
9 & Baseline-parity replication & Closed & Cross-model and cross-seed \\
10 & Cache-miss-to-CUA fallback (Muscle-Mem gap) & \textbf{Closed $n{=}4$} & WebArena warm-$\Delta$ $-4.0 \to -1.0$, matches Muscle-Mem $-0.75$ (Welch's 2-sided $p \approx 0.84$) \\
11 & Compile-LLM robustness (Gemini vs Claude) & \textbf{Closed $n{=}3$} & WebArena cold $6.0 \approx 6.33$; rejection $83 \approx 89\%$; mechanism shared \\
12 & Embedding-selector robustness (MiniLM vs bge-large) & \textbf{Closed} & Both $100\%$ functional retrieval; bge-large wider safe $\tau$ window \\
\bottomrule
\end{tabular}
\end{table}

\end{document}